\documentclass[journal]{IEEEtran}
\usepackage{amsmath,amsfonts}
\usepackage{array}
\usepackage{textcomp}
\usepackage{stfloats}
\usepackage{url}
\usepackage{verbatim}
\usepackage{graphicx}
\usepackage{amssymb}
\usepackage{hyperref}
\usepackage{graphicx}
\usepackage{amsmath}
\usepackage{longtable}
\usepackage{algorithm} 
\usepackage{algpseudocode} 
\usepackage{mathrsfs}
\usepackage{subcaption}
\usepackage{mathtools}
\usepackage{pifont}
\usepackage{color}
\usepackage{lineno}
\usepackage{makecell} 
\usepackage{pdflscape}
\usepackage{adjustbox}
\usepackage[utf8]{inputenc}
\usepackage{tabularx}
\usepackage{blindtext}
\usepackage{longtable}
\usepackage{lscape}
\usepackage{setspace}
\usepackage{notoccite} 
\usepackage{lscape} 
\usepackage{mwe}
\usepackage{booktabs}
\usepackage{amsthm}

\theoremstyle{definition}

\theoremstyle{definition}

\newcommand{\RNum}[1]{\lowercase\expandafter{\romannumeral #1\relax}}
\newcommand{\RNumU}[1]{\uppercase\expandafter{\romannumeral #1\relax}}
\usepackage[numbers]{natbib}
\def\BibTeX{{\rm B\kern-.05em{\sc i\kern-.025em b}\kern-.08em
    T\kern-.1667em\lower.7ex\hbox{E}\kern-.125emX}}
\usepackage{balance}
\begin{document}
\title{Flexi-Fuzz least squares SVM for Alzheimer's diagnosis: Tackling noise, outliers, and class imbalance}
\author{Mushir Akhtar, \IEEEmembership{Graduate Student Member,~IEEE},  A. Quadir, \IEEEmembership{Graduate Student Member,~IEEE}, M. Tanveer{$^*$}, \IEEEmembership{Senior Member,~IEEE}, Mohd. Arshad, for the Alzheimer’s Disease Neuroimaging Initiative{$^{**}$}
\thanks{ \noindent $^*$Corresponding Author\\
    Mushir Akhtar,  A. Quadir, M. Tanveer and Mohd. Arshad are with the Department of Mathematics, Indian Institute of Technology Indore, Simrol, Indore, 453552, India (e-mail: phd2101241004@iiti.ac.in, mscphd2207141002@iiti.ac.in, mtanveer@iiti.ac.in, arshad@iiti.ac.in).\\  $^{**}$Data utilized in this article were sourced from the Alzheimer’s Disease Neuroimaging Initiative (ADNI) database (adni.loni.usc.edu). While the investigators within ADNI contributed to the design and implementation of ADNI and/or provided data, they were not involved in the analysis or writing of this paper.
 }}
\maketitle
\begin{abstract}
Alzheimer’s disease (AD) is a leading neurodegenerative condition and the primary cause of dementia, characterized by progressive cognitive decline and memory loss. Its progression, marked by shrinkage in the cerebral cortex, is irreversible.
Numerous machine learning algorithms have been proposed for the early diagnosis of AD. However, they often struggle with the issues of noise, outliers, and class imbalance. To tackle the aforementioned limitations, in this article, we introduce a novel, robust, and flexible membership scheme called Flexi-Fuzz. This scheme integrates a novel flexible weighting mechanism, class probability, and imbalance ratio. The proposed flexible weighting mechanism assigns the maximum weight to samples within a specific proximity to the center, with a gradual decrease in weight beyond a certain threshold. This approach ensures that samples near the class boundary still receive significant weight, maintaining their influence in the classification process. Class probability is used to mitigate the impact of noisy samples, while the imbalance ratio addresses class imbalance. Leveraging this, we incorporate the proposed Flexi-Fuzz membership scheme into the least squares support vector machines (LSSVM) framework, resulting in a robust and flexible model termed Flexi-Fuzz-LSSVM. We determine the class-center using two methods: the conventional mean approach and an innovative median approach, leading to two model variants, Flexi-Fuzz-LSSVM-I and Flexi-Fuzz-LSSVM-II. By utilizing the median's robustness, Flexi-Fuzz-LSSVM-II effectively handles outliers and non-symmetrical distributions. To validate the effectiveness of the proposed Flexi-Fuzz-LSSVM models, we evaluated them on benchmark UCI and KEEL datasets, both with and without label noise. Additionally, we tested the models on the Alzheimer’s Disease Neuroimaging Initiative (ADNI) dataset for AD diagnosis.
Experimental results demonstrate the superiority of the Flexi-Fuzz-LSSVM models over baseline models, highlighting their potential in improving AD diagnosis and handling noisy data.
\end{abstract}

\section{Introduction}
\IEEEPARstart{A}{lzheimer’s disease (AD)} is a chronic and progressive neurodegenerative condition \cite{zhang2020survey, benussi2022classification}, accounting for approximately $70\%$ of dementia cases \cite{khojaste2022deep}. It primarily affects the brain's cognitive functions, leading to disorientation, mood disturbances, and severe cognitive decline \cite{sampathkumar2021adiag, sun2021improved}. The prevalence of AD is projected to increase dramatically, with estimates suggesting that by $2050$, one in every $85$ individuals will be affected \cite{porsteinsson2021diagnosis}.
Currently, about $6.7$ million individuals aged $65$ and above are living with AD, making it the sixth-leading cause of mortality in the United States. The socioeconomic impact of AD is profound, encompassing significant healthcare costs, social welfare provisions, and substantial income losses for families, amounting to an estimated $345$ billion in the United States in $2023$ \cite{AD2023Report}.
Numerous studies indicate that early detection and intervention can significantly slow the progression of AD. Therefore, to mitigate further growth, treatment should commence at the earliest detectable stage. Recently, there has been a burgeoning interest in leveraging machine learning algorithms for AD diagnosis. The application of machine learning in AD detection has yielded promising outcomes and represents a vibrant area of research in the past decade \cite{richhariya2020diagnosis, malik2022alzheimer}. 
For a detailed overview of machine learning models for AD diagnosis, we refer interested readers to \cite{tanveer2020machine}.

Support vector machine (SVM) \cite{cortes1995support} is one of the prominent machine learning techniques widely used for classifying AD \cite{richhariya2020diagnosis} and other diseases \cite{kumari2024diagnosis}. Unlike numerous classification methodologies, SVM is distinguished by its strong mathematical foundation. It employs the principle of structural risk minimization (SRM) for regularization, simultaneously minimizing the empirical risk. This approach results in strong generalization performance with reduced overfitting of the data \cite{vapnik1999nature}. Moreover, the optimization problem of SVM produces a unique optimal solution, unlike techniques such as artificial neural networks (ANN), which are susceptible to the issue of local minimum \cite{gori1992problem}. Despite these strengths, SVM is constrained by its high computational complexity \cite{abdul2024granular}.

To subsidize the overhead of computational complexity of SVM, \citet{suykens1999least} introduced a least squares variant of SVM known as the least squares support vector machine (LSSVM). Unlike SVM, LSSVM employs a quadratic loss function instead of the hinge loss function, allowing for the inclusion of equality constraints in the classification problem. Consequently, the solution can be obtained directly by solving a set of linear equations, bypassing the need for quadratic programming. The computation time of LSSVM is significantly less compared to SVM. However, it is worth noting that both SVM and LSSVM are susceptible to the effects of noise and outliers \cite{akhtar2024advancing, akhtar2023roboss}, which are ubiquitous in AD diagnosis. Outliers are data points that deviate significantly from the majority of the data, while noise refers to random fluctuations or errors in the data \cite{smiti2020critical}. The presence of outliers and noise in the dataset can distort the learning process and lead to poor generalization performance of the model. 
\par
To ameliorate the repercussions of noise and outliers, fuzzy theory has been successfully applied in machine learning models \cite{wang2005new, sajid2024neuro}.
Fuzzy theory uses the distance between an instance and its corresponding class-center to generate a membership function. This membership function allows the model to assign a degree of belongingness to each data point, enabling it to effectively handle outliers and noise in the dataset. This approach enhances the robustness of the model and improves its ability to generalize to unseen data. Class-imbalance learning (CIL) in the existence of noise and outliers is another major task in classification problems, specifically in the biomedical domain. Imbalanced datasets, characterized by an unequal distribution of samples across classes, often lead to poor performance on minority classes due to overfitting, resulting in irretrievable loss \cite{rezvani2023broad}. Generally, there are two primary strategies to tackle the class-imbalance problem: resampling and re-weighting. Resampling is a data-level approach that involves manipulating the dataset by either oversampling the under-represented (minority) class instances or undersampling the frequent (majority) class instances \cite{chawla2002smote}. However, the application of resampling has some drawbacks. Specifically, oversampling may lead to overfitting and reduced training efficiency due to an increase in redundant samples, while undersampling can result in information loss and suboptimal performance on frequent class. On the other hand, re-weighting is an algorithmic-level approach that entails redesigning the fuzzy scheme to compel the model to prioritize attention to the more challenging under-represented instances. The aforementioned issues are prevalent in various real-world domains, including medical diagnosis, particularly in AD diagnosis \cite{10561527, tanveer2024ensemble}.

To enhance the performance of CIL in the presence of noise and outliers, several SVM-based algorithms have been proposed. These variants rely on fuzzy membership schemes to mitigate the impact of noise and outliers on the classification method. Fuzzy support vector machine (FSVM) \cite{lin2002fuzzy} handles the noise/outliers issue by assigning fuzzy membership values to the input data points based on their distance from the class-center. The crucial aspect of FSVM variants lies in how they allocate membership value to the input sample. In \cite{zhou2009fuzzy}, an improved dual fuzzy membership function is introduced, in which data points exhibit varying degrees of membership in different regions. Despite the effectiveness of fuzzy concepts in mitigating the effects of noise/outliers, FSVM variants still encounter challenges in class-imbalance problems.
To tackle these challenges, Batuwita and Palade \cite{batuwita2010fsvm} introduced fuzzy support vector machine for CIL (FSVM-CIL) with four distinct membership schemes, comprising two center-based and two hyperplane-based approaches. The hyperplane-based membership scheme presumes that the initially derived hyperplane always accurately predicts the final hyperplane, which is not always accurate. Fan et al. \cite{fan2017entropy} introduced entropy-based FSVM (EFSVM), which considers the class certainty of the samples but struggles with border samples and outliers. To further overcome these limitations in CIL with noise and outliers, Tao et al. \cite{tao2020affinity} proposed affinity and class probability-based fuzzy support vector machine (ACFSVM). It utilizes class probability and support vector data description (SVDD) \cite{tax2004support} to compute the fuzzified value of the data samples. However, the calculation of affinity significantly influences the computation of membership value, leading to a notable increase in computational complexity. The detailed limitations of the existing fuzzy schemes are discussed in Section \ref{Limitation of existing schemes}.

Taking inspiration from prior research endeavors, in this paper, we introduce a novel, robust, and flexible membership scheme, named Flexi-Fuzz, meticulously engineered to tackle the pervasive challenges of noise, outliers, and class imbalance, which are critical in the diagnosis of AD. The core of the Flexi-Fuzz scheme is its robust and flexible weighting mechanism. By calculating the distance between each sample and the class-center, it assigns weights that reflect the proximity of the samples. Maximum weight is assigned to samples within a specified neighborhood of the center, with a gradual decrease beyond that threshold, ensuring that samples near the class boundary retain significant influence. To overcome the influence of class noise, we integrate class probability, which involves adjusting the weights based on the likelihood of a sample belonging to a particular class, thus reducing the influence of noisy samples that could otherwise distort the learning process. Finally, to address class imbalance, we amalgamate an imbalance ratio into the weighting mechanism, ensuring that minority-class samples are adequately represented during training, thus improving model performance on imbalanced datasets. Subsequently, by incorporating the proposed Flexi-Fuzz membership scheme into the framework of LSSVM, we propose a robust and flexible LSSVM model named Flexi-Fuzz-LSSVM.  To enhance the versatility of our approach, we determine the class-center using two methods: the conventional mean approach and an innovative median approach. This leads to the development of two model variants, Flexi-Fuzz-LSSVM-I and Flexi-Fuzz-LSSVM-II, each producing distinct sets of fuzzified samples. The key contributions of this work can be encapsulated as follows:
\begin{enumerate}  
\item We present Flexi-Fuzz, a novel, robust, and flexible membership scheme designed to effectively combat the challenges posed by noise, outliers, and class imbalance. This scheme allocates maximum weight to samples within a specified proximity of the center, gradually reducing the weight beyond this threshold, thereby allowing for tailored flexibility. Additionally, class probability and imbalance ratio are employed to attenuate the effects of noise and rectify class imbalance, respectively.
\item We amalgamate the proposed Flexi-Fuzz membership scheme into the LSSVM and present a robust and flexible least squares support vector machine (Flexi-Fuzz-LSSVM) for efficiently tackling the issue of CIL in the presence of noise and outliers. By leveraging the LSSVM framework, the proposed method eliminates the need to solve the computationally expensive quadratic programming problem. Instead, the solution is directly obtained by solving a set of linear equations.
\item We augment our contribution by utilizing two distinct methodologies for determining the class-center: the conventional mean approach and the innovative median approach. Consequently, we develop two model variants, Flexi-Fuzz-LSSVM-I and Flexi-Fuzz-LSSVM-II, each producing a unique set of fuzzified samples.
\item We evaluate the effectiveness of the proposed Flexi-Fuzz-LSSVM models on UCI and KEEL datasets (with and without label noise). The empirical results illustrate that the proposed Flexi-Fuzz-LSSVM models exhibit superior performance compared to numerous baseline models.
\item To validate the effectiveness of the proposed Flexi-Fuzz-LSSVM models in the detection of AD, we conducted experiments using the Alzheimer’s Disease Neuroimaging Initiative (ADNI) dataset. The empirical investigations provide compelling evidence of the proposed Flexi-Fuzz-LSSVM models applicability in the early diagnosis of AD.
\end{enumerate}
The rest of this work is structured as follows: Section \ref{Related work} provides an overview of the existing fuzzy schemes. Section \ref{Proposed work} presents the limitations of existing fuzzy schemes and introduces the proposed Flexi-Fuzz membership scheme. Additionally, in Section \ref{Proposed work}, we derive the mathematical formulation of the proposed Flexi-Fuzz-LSSVM model. Section \ref{Experimental results} presents experimental results and statistical comparisons on the UCI and KEEL datasets, as well as on the ADNI dataset. Finally, Section \ref{Conclusion} concludes the paper with future directions.
\section{Related Work}\label{Related work}
\subsection{Existing Fuzzy Schemes}
In various real-world scenarios, machine learning algorithms are tasked with learning to distinguish between two distinct classes of input points. However, not all training points carry the same level of importance. Some points are considered more meaningful in accurately classifying the data, while others, such as noise or irrelevant data, are less significant. That is, each training point $x_i$ is linked to a fuzzy membership $m_i$, which serves as an indicator of the degree to which the point is inclined or relevant to a specific class in the classification problem. Assigning membership values to training samples manually by leveraging one's intuition and understanding of the data can prove effective. Nevertheless, in many cases, the feasibility of this approach diminishes, particularly when confronted with complex or large datasets. To overcome this challenge, general purpose membership functions (GPMFs) can serve as a viable solution \cite{sevakula2017compounding}. GPMFs are membership functions that are based on commonly observed patterns and structural understanding of data, making them applicable to a wide range of datasets. Here, we discuss some existing GPMFs, which are based on the distance from the class-center and hyperplane. The GPMFs defined in prior research \cite{lin2002fuzzy, batuwita2010fsvm} are provided as follows:
\begin{align}
& m_{\text {lin }}^{\text {cen }}\left(x_i\right)=1-\frac{d_i^{\text {cen }}}{r+\delta}, \label{membership_function_1}\\
& m_{\exp }^{\text {cen }}\left(x_i\right)=\frac{2}{1+\exp \left(\gamma d_i^{\text {cen }}\right)}, \quad \gamma \in[0,1], \label{membership_function_2}\\
& m_{\text {lin }}^{\text {hyp }}\left(x_i\right)=1-\frac{d_i^{\text {hyp }}}{\max \left(d_i^{\text {hyp }}\right)+\delta}, \label{membership_function_3}
\\
& m_{\exp }^{\text {hyp }}\left(x_i\right)=\frac{2}{1+\exp \left(\gamma d_i^{\text {hyp }}\right)}, \quad \gamma \in[0,1] \label{membership_function_4}.
\end{align}
Here, $m(x_i)$ represents the assigned membership value for the $i^{th}$ training sample, with its values spanning from $0$ to $1$. The parameter $d_i^{\text{cen}}$ signifies the distance of the $i^{th}$ training sample to its respective class-center, where the class-center is calculated by computing the mean of all samples within the class. The variable $r$ denotes the class radius, defined as $r = \max (d_i^{\text{cen}})$. Hence, the design of $m_{\text {lin}}^{\text{cen}}$ is premised on the concept that the belongingness of a sample to the class diminishes linearly as the sample moves farther from the class-center. $m_{\text {exp}}^{\text{cen}}$ is formulated based on the same principle as $m_{\text {lin}}^{\text{cen}}$, with the sole distinction that the membership value in this case decays exponentially as the sample moves farther from the class-center. In the cases of $m_{\text {lin}}^{\text{hyp}}$ and $m_{\text{exp}}^{\text{hyp}}$, a separating hyperplane is initially constructed using classical SVM techniques. Here, $d_i^{\text{hyp}}$  represents the absolute value of the functional margin of the $i^{th}$ training sample relative to the initial separation hyperplane. The rationale behind these membership values is grounded in the idea that training samples in closer proximity to the separation hyperplane convey more information and, consequently, merit higher membership values compared to others. The value of $m_{\text {lin}}^{\text{hyp}}$ decays linearly as $d_i^{\text{hyp}}$ increases, while the value of $m_{\text {exp}}^{\text{hyp}}$ decays exponentially. One significant drawback of hyperplane-based membership is its assumption that the initially derived hyperplane always accurately predicts the final hyperplane. However, this assumption may not always hold true in practice. $\delta$ represents a small positive value, ensuring that $m(x_i)$ consistently maintains a value greater than zero, while $\gamma$ serves the purpose of governing the degree to which exponential decay influences membership values. The aforementioned membership functions perform on par when dealing with outliers; however, they still have limitations. Notably, these functions may prove less effective in addressing challenges associated with CIL. To overcome this limitation, \citet{tao2020affinity} proposed an affinity and class probability-based fuzzy membership function. It uses the idea of SVDD and calculates the affinity of the predominant class, thereby pinpointing outlier samples located within the majority-class. The expression to determine the affinity of a sample is as follows:
\begin{align}
\varphi_{aff}\left(x_i\right)= \begin{cases}0.8\left(\frac{d_i^{s v d d}-\min \left(d_i^{s v d d}\right)}{\max \left(d_i^{s v d d}\right)-\min \left(d_i^{s v d d}\right)}\right), &  d_i^{s v d d}<\beta \times R, \\ 0.2\left(\exp \left(\rho\left(1-\frac{d_i^{s v d d}}{\beta \times R}\right)\right)\right), &  d_i^{s v d d} \geq \beta \times R.\end{cases}
\end{align}
Here, $\beta \in  \left(0, 1\right]$ serves as a trade-off parameter influencing the size of potential outliers and border samples, $\rho > 0$ is a decay parameter, and $d_i^{svdd}$ represents the distance of $i^{th}$ sample ($x_i$) from the class-center. The value of $R$ can be calculated as the mean distance of all support vectors situated on the boundary from the center of the hypersphere.\\
Furthermore, the approach utilizes the $k$-nearest neighborhood technique to ascertain the class probability associated with samples within the predominant class. The probability value, denoted as $p(x_i)$, for the sample $x_i$ is computed by determining by the ratio of same-class data points within the $k$ nearest neighborhood data points (selected based on proximity) to the total number of neighbors $k$. Finally, fuzzy membership based on affinity and class probability is calculated as follows:
\begin{align} \label{Aff class prob membership}
m_{aff}\left(x_i\right)= \begin{cases}1, & x_i \in \text {minority-class} \\ \varphi_{aff}\left(x_i\right) \times p\left(x_i\right), &  x_i \in \text { majority-class.}\end{cases}
\end{align}
This approach proves efficient when dealing with outliers and class imbalance issues; however, it still has some limitations. One notable drawback is the high computational complexity associated with the calculation of affinity, which necessitates solving the QPP inherent in the SVDD technique.
\section{Proposed Work} \label{Proposed work}
In this section, we undertake an analysis of the constraints present in the existing fuzzy schemes. To address their demerits, we introduced a novel, robust, and flexible fuzzy scheme, termed Flexi-Fuzz, specifically designed to address the challenges of noise, outliers, and imbalanced data prevalent in AD diagnosis. By integrating the proposed Flexi-Fuzz membership scheme into the framework of LSSVM, we introduce advanced, robust, and flexible fuzzy LSSVM models for CIL, referred to as Flexi-Fuzz-LSSVM.
\subsection{Limitations of Existing Fuzzy Schemes} \label{Limitation of existing schemes}
Though the membership functions stipulated in equations (\ref{membership_function_1})-(\ref{membership_function_4}) and (\ref{Aff class prob membership}) demonstrate  resistance against noise and outliers, they have certain limitations, including:
\begin{enumerate}
\item Membership function, namely $m_{\text{lin}}^{\text{cen}}$, demonstrates a notable reliance on the distance from the class-center. In scenarios like Figure \ref{fig:first},  where the data distribution results in the majority of training samples being situated on the class boundary, the assignment of membership to most data points approaches to zero, as per equation (\ref{membership_function_1}). Consequently, the influence of these types of samples is diminished on the classification process.
\item The functions $m_{\text{lin}}^{\text{cen}}$ and $m_{\text{exp}}^{\text{cen}}$ exhibit a linear and exponential decrease, respectively, as the distance of the input point from its class-center increases.
Despite this, both functions encounter a significant limitation in appropriately representing the importance of data points situated in close proximity to the class-center. These points should merit equal consideration, as they share comparable significance with the class-center. However, the existing schemes, $m_{\text{lin}}^{\text{cen}}$ and $m_{\text{exp}}^{\text{cen}}$, assign lower membership values to these samples compared to those at the class-center, thereby reducing their influence on the classification process. 
\item The membership functions (\ref{membership_function_1})-(\ref{membership_function_4}) solely depend on the distance from the class-center or initial hyperplane. Nonetheless, training samples with equal distances may exhibit varying influences on the formation of the classifier. For example, in scenarios like Figure \ref{fig:third}, the class noise, despite having the same distance as the normal training sample, should yield a lower membership value. This observation highlights that considering only distance-based membership schemes is insufficient for accurately characterizing class membership.
 \begin{figure*}[ht]
\begin{minipage}{.333\linewidth}
\centering
\subfloat[]{\label{fig:first}\includegraphics[scale=0.40]{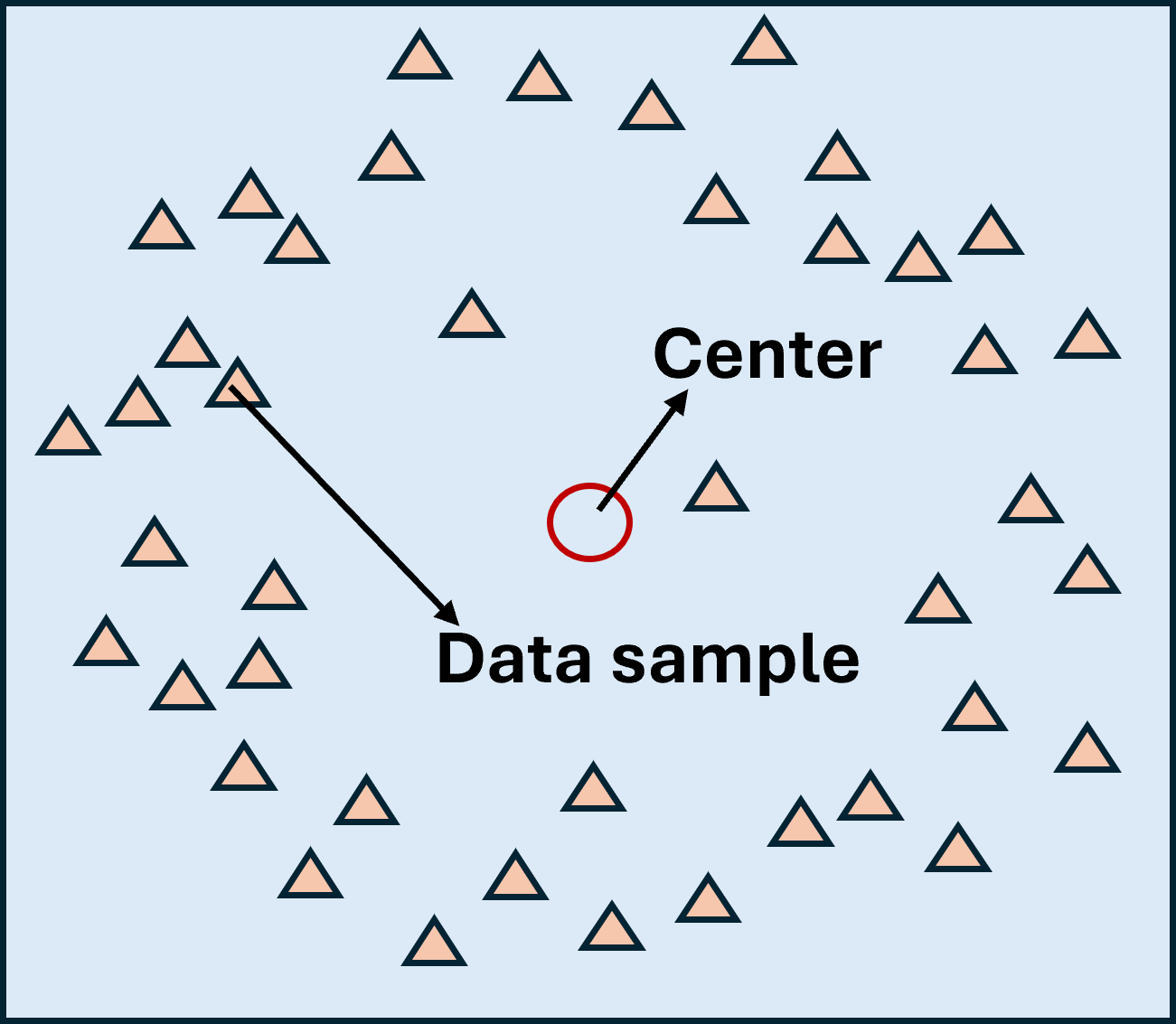}}
\end{minipage}
\begin{minipage}{.333\linewidth}
\centering
\subfloat[]{\label{fig:second}\includegraphics[scale=0.40]{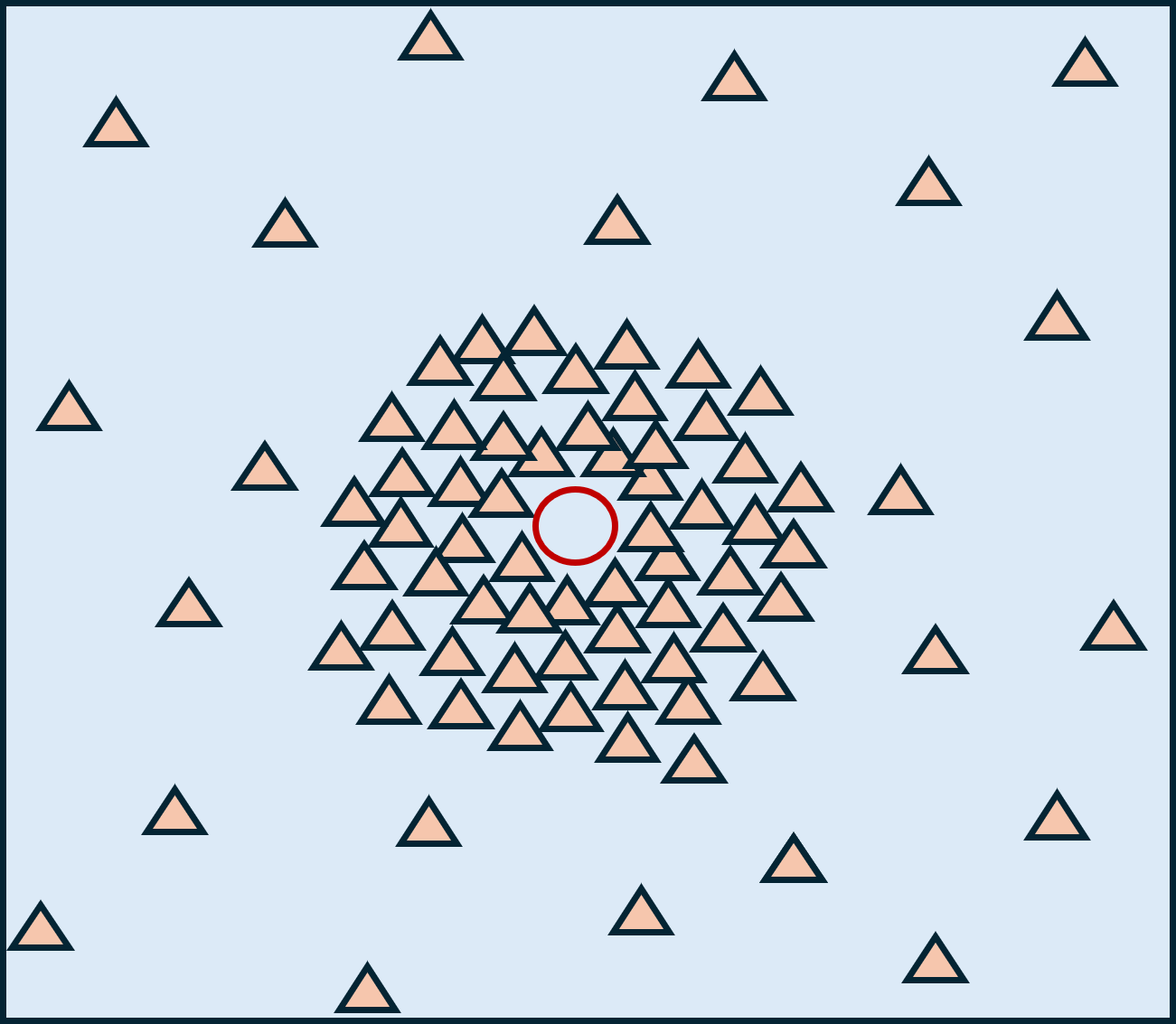}}
\end{minipage}
\begin{minipage}{.333\linewidth}
\centering
\subfloat[]{\label{fig:third}\includegraphics[scale=0.40]{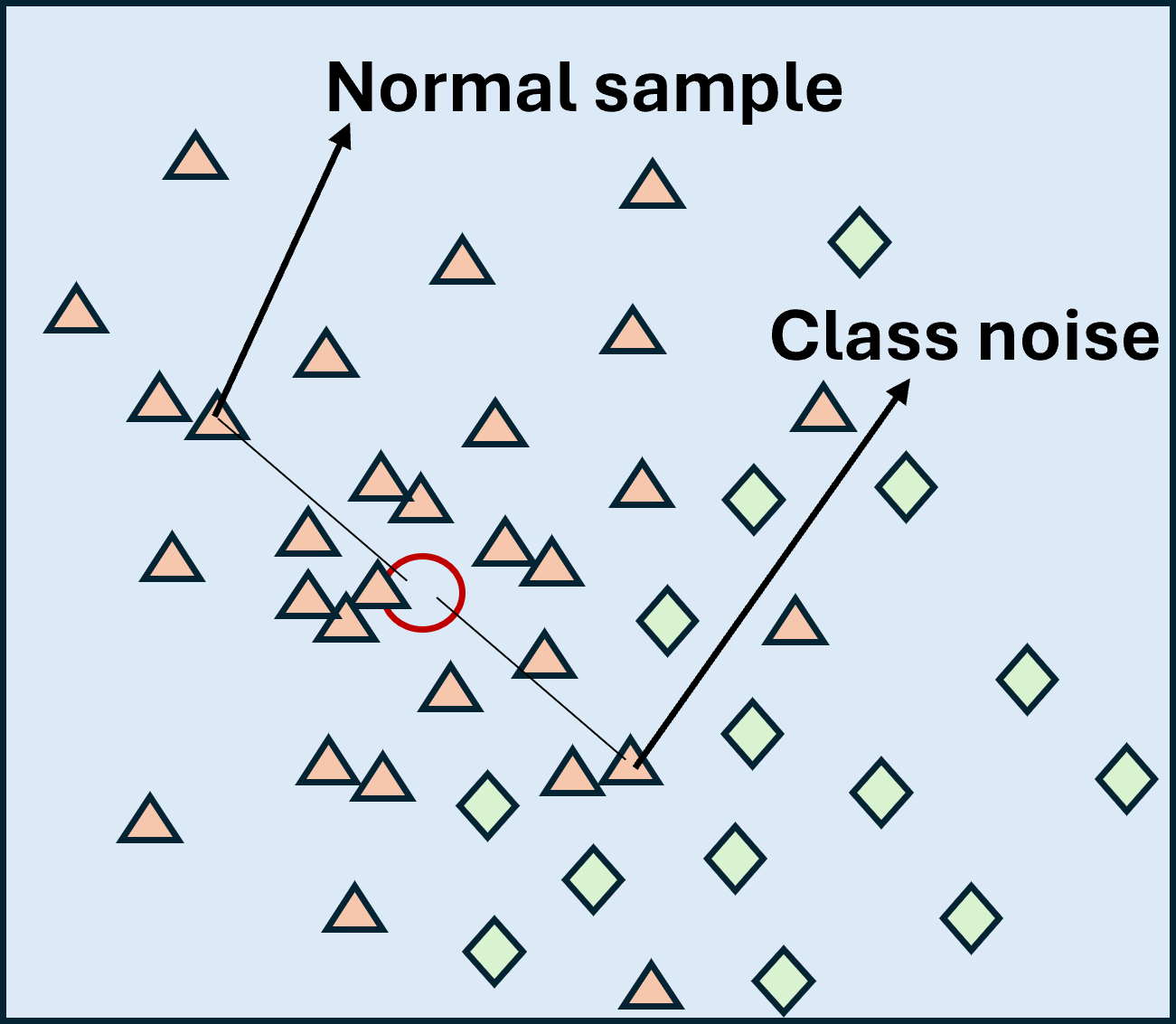}}
\end{minipage}
\caption{(a) Depicts a sample distribution where the majority of samples are dispersed farther from the class-center. (b) Illustrates a sample distribution with the majority of samples densely clustered around the class-center. (c) Demonstrates that noise samples are positioned at similar distances from the class-center as the normal samples, highlighting the challenge of distinguishing noise from legitimate data points based solely on their proximity to the center.}
\label{gb-generation}
\end{figure*}
\item Membership functions based on the class-center are highly dependent on the method used to calculate the center. Traditionally, these functions determine the class-center by computing the mean of all samples associated with that class. However, the mean is not a robust estimator, especially in the presence of outliers or non-symmetrical data \cite{crow1967robust}. Mathematically, the mean is highly susceptible to extreme values, which can significantly distort its accuracy. Outliers located far from the majority of samples can skew the mean, leading to a compromised estimation of the class-center. This sensitivity undermines the robustness of the center calculation, necessitating alternative methods that are less influenced by outliers to provide a more reliable characterization of the class-center.
\item The underlying principle embedded in $m_{\text{lin}}^{\text{hyp}}$ and $m_{\text{exp}}^{\text{hyp}}$ posits that training samples that are close to the initial hyperplane contain more useful information than those that are farther away. Consequently, higher membership values are assigned to training samples closer to the initial hyperplane, reflecting their perceived greater relevance. However, it is crucial to note that these two membership functions operate under the assumption that the initial hyperplane learned is potentially unaffected by outliers and is a correct estimate of the final hyperplane. Unfortunately, these assumptions fail in many instances.  Furthermore, the procedure leading to the determination of the initial hyperplane is computationally expensive. 
\end{enumerate}
\subsection{Proposed Flexi-Fuzz Membership Scheme}
To address the aforementioned limitations, we have developed an innovative and robust membership scheme, termed Flexi-Fuzz. The proposed Flexi-Fuzz scheme is meticulously structured around three pivotal steps: (1) calculating the weight for each training sample through a novel flexible weighting mechanism; (2) determining the class probability to refine the weighting process; and (3) assessing the imbalance ratio to ensure equitable representation of all classes.
\subsubsection{Proposed flexible weighting mechanism}
\begin{figure}[htp]
    \centering   \includegraphics[width=0.40\textwidth,keepaspectratio]{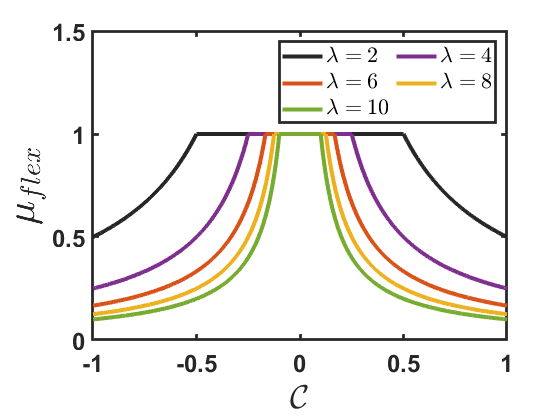}
    \caption{Illustrates the proposed flexible weighting mechanism, showing the effect of varying the flexible parameter \(\lambda\). Different \(\lambda\) values adjust the weighting of samples based on their distance from the class-center, with higher weights assigned to central samples and gradually decreasing weights for more distant samples.}
    \label{fig:proposed_weighting_scheme}
\end{figure}

In traditional methodologies, sample weight decreases linearly or exponentially as the distance from the class-center increases. This results in the maximum weight being assigned only to samples at the center, while those near the boundary receive negligible weights. Our novel approach challenges this paradigm by advocating for maximum weight to be assigned to samples within a specified neighborhood of the center, with a gradual decrease beyond that threshold. Additionally, our approach ensures that samples near the boundary do not receive negligible weights, thereby ensuring their influence is not entirely neglected in the classification process. The mathematical articulation of the proposed flexible weighting scheme is outlined as follows:
\begin{align} \label{proposed_weight}
\mu_{flex}\left(x_i\right)=
\begin{cases} 1, & |x_i - \mathcal{C}| <\frac{\mathcal{R}}{\lambda} \\
\frac{\mathcal{R}}{\lambda \times |x_i - \mathcal{C}|}, &  |x_i - \mathcal{C}| \geq \frac{\mathcal{R}}{\lambda},\end{cases}
\end{align}
where $\mathcal{C}$ and $\mathcal{R}$ denote the center and radius of the corresponding class, respectively, and $\lambda \geq 1$ is the flexible parameter. For $\lambda =1$, the weighting scheme uniformly assigns a maximum weight of $1$ to all samples. For $\lambda =2$, it assigns a maximum weight of $1$ exclusively to samples situated within the radius of $\frac{\mathcal{R}}{2}$, and so on. This exemplifies the parameter $\lambda$ as a key factor influencing the weighting scheme, allowing for tailored flexibility. Thus, by adjusting $\lambda$, the scheme can be fine-tuned to vary the influence of samples based on their proximity to the class-center. This flexibility allows for more precise control over the classification process, enhancing the model's adaptability to different datasets and noise levels. The proposed flexible weighting scheme, illustrated in Figure \ref{fig:proposed_weighting_scheme}, visually demonstrates the impact of different values of $\lambda$. The ability to manipulate $\lambda$ enables the customization of emphasis on proximity to the class-center. This is particularly useful in applications where the significance of boundary samples varies, allowing for optimized performance across diverse scenarios.
\subsubsection{Determining the class probability}
Data points with equal distances from the class-center may exhibit distinct contributions to the classifier's formation, indicating that relying solely on weight values determined by distance is insufficient to accurately characterize their belongingness to a specific class. To address this limitation and enhance performance, class probability values are introduced. This inclusion aims to reduce the influence of class noise in the training phase. Specifically, the class probability for each data input is precisely determined as the ratio of points belonging to the same class within a predefined fixed neighborhood to the total number of points within that neighborhood. To elaborate further, for a given training sample $x_i$, the procedure involves the selection of its $k$-nearest neighbors, denoted as $\left\{x_{i 1}, x_{i 2}, \ldots, x_{i k}\right\}$. Then, the count of samples within this selected set that belong to the same class as $x_i$ $(n^{own}_i)$ is determined. Finally, the class probability for the sample $x_i$ is computed as
\vspace{-0.3cm}
\begin{align} \label{probability_value}
p(x_i)=\frac{n^{own}_i}{k}.
\end{align}
The larger value of $p(x_i)$ signifies a higher probability for the sample $x_i$ to belong to its own class. Consequently, a lower probability value for noise reduces its impact on the construction of hyperplanes. This observation underscores the utility of class probability in enhancing the robustness of the hyperplane construction process, contributing to a more reliable and accurate classification model.
\subsubsection{Evaluating the imbalance ratio}
To mitigate the challenges posed by imbalanced class distributions, we utilized the imbalance ratio for updating the weights. Without loss of generality, we assume that the positive class represents the minority-class, while the negative class represents the majority-class. The imbalance ratio $(I.R.)$ is then computed as the ratio of the number of majority samples to the number of minority samples. Let $\mu_{flex}(x_i^{+})$ and $\mu_{flex}(x_i^{-})$ denote the weight values of a positive-class example $x_i^{+}$ and a negative-class example $x_i^{-}$, respectively. Subsequently, the weights assigned to each sample are updated as follows:
\begin{align}\label{updatedscheme1}
\mu_{flex}^{'}(x_i^{+}) =& \mu_{flex}(x_i^{+}) \times I.R.,
\end{align}
\begin{align}\label{updatedscheme2}
\mu_{flex}^{'}(x_i^{-}) =& \mu_{flex}(x_i^{-}) \times \frac{1}{I.R.}.   
\end{align}
Here, $\mu_{flex}(x_i)$ takes a value in between $\left(0,1\right]$. As a result, a positive-class sample can have a membership value ranging from $0$ to $I.R.$, whereas a negative-class sample can have a membership value ranging from $0$ to $\frac{1}{I.R.}$. This adaptation serves to address the imbalanced class distribution by scaling the membership values accordingly.

Finally, combining the updated weighting scheme, as defined in equations (\ref{updatedscheme1}) and (\ref{updatedscheme2}), with the class probability value calculated using equation (\ref{probability_value}), we define the proposed Flexi-Fuzz membership scheme as follows:
\begin{align} \label{final proposed scheme}
\mathcal{M}_{flexi}^{fuz}\left(x_i\right)= \begin{cases} \mu_{flex}(x_i) \times p\left(x_i\right) \times I.R., & x_i \in \text {minority-class} \\ \mu_{flex}(x_i) \times p\left(x_i\right) \times \frac{1}{I.R.}, &  x_i \in \text { majority-class.}\end{cases}
\end{align}
This formulation incorporates the proposed flexible weighting mechanism with both the class probability and imbalance ratio, offering a comprehensive and nuanced representation of sample belongingness within the Flexi-Fuzz membership scheme. This innovation ensures that the Flexi-Fuzz scheme remains robust, adaptive, and highly effective in managing complex classification tasks, particularly in the context of AD diagnosis, where noise, outliers, and class imbalance are common challenges. The method to compute the proposed Flexi-Fuzz membership scheme is briefly described
in Algorithm \ref{algorithm1}.
\begin{algorithm}
 \caption{Algorithm for the proposed Flexi-Fuzz membership scheme}
 \label{algorithm1}
 \textbf{Input:}\\
 The training dataset:  $\left\{x_i,y_i\right\}_{i=1}^n$, $y_i \in\{-1,1\}$;\\
 The parameters: flexible parameter $\lambda$, $k$-nearest neighbor parameter $k$;\\
 \textbf{Output:}\\ 
 The Flexi-Fuzz membership value for each sample;\\
1: Compute the flexible weighting scheme using equation (\ref{proposed_weight});\\
2: Compute the class probability using equation (\ref{probability_value});\\
3: Compute the updated weighting scheme using equations (\ref{updatedscheme1}) and (\ref{updatedscheme2});\\
4: Determine the proposed Flexi-Fuzz membership scheme using equation (\ref{final proposed scheme}). \\
 \end{algorithm}
\subsection{Computational Complexity}
Let a binary classification task have a total of $n$ data points, and $n_1$ and $n_2$ are the counts of positive and negative data points, respectively. The computational complexity of the proposed Flexi-Fuzz membership scheme (equation \eqref{final proposed scheme}) can be broken down into three main components. Firstly, it involves calculating the flexible weighting scheme value. This computation includes determining the class-center and radius, computing the distance between each class-center and sample, and evaluating the flexible weighting scheme value for each sample using equation (\ref{proposed_weight}). This step has a time complexity of $\mathcal{O}(1) + \mathcal{O}(1) + \mathcal{O}(n)$. Secondly, it needs to calculate the class probability according to equation (\ref{probability_value}). This calculation involves iterating over $n_1$ and $n_2$ samples, leading to a time complexity of $\mathcal{O}(n_1) + \mathcal{O}(n_2)$ \cite{rezvani2019intuitionistic}. Lastly, it requires to compute the imbalance ratio, which has a fixed time complexity of $\mathcal{O}(1)$. Therefore, the overall computational complexity of the Flexi-Fuzz membership scheme is approximately $\mathcal{O}(n) + \mathcal{O}(n_1) + \mathcal{O}(n_2)$.

\subsection{Proposed Flexi-Fuzz-LSSVM}
Through the integration of the proposed Flexi-Fuzz membership scheme into the framework of LSSVM, we propose a novel robust classifier for class-imbalance problems named Flexi-Fuzz-LSSVM. The objective function of the proposed Flexi-Fuzz-LSSVM is formally defined as follows:
\begin{align} \label{ProposedLSSVM}
\min _{w, b} \hspace{0.2cm} &\frac{1}{2}\|w\|^2+\frac{1}{2} C \sum_{i=1}^n \mathcal{M}_i \xi_i^2 \nonumber\\
\text {s.t.} \hspace{0.2cm} &y_i\left(w^{\top} \phi\left(x_i\right)+b\right)=1-\xi_i, \quad i=1,2, \ldots,n,
\end{align}
where $C > 0$ is the regularization term and $\mathcal{M}_i$ denotes the Flexi-Fuzz membership value. As the dimensionality of $w$ becomes infinite, solving the primal problem (\ref{ProposedLSSVM}) directly is not feasible. Consequently, we proceed by formulating the corresponding Lagrangian:
\begin{align} \label{Proposed_lagrange_LSSVM}
\mathcal{L}=&\frac{1}{2}\|w\|^2+\frac{1}{2} C \sum_{i=1}^n \mathcal{M}_i \xi_i^2 -\sum_{i=1}^n \alpha_i\left\{y_i\left[w^{\top} \phi\left(x_i\right)+b\right]-1+\xi_i\right\},
\end{align}
where $\alpha_i$'s are the Lagrangian multipliers. The optimal conditions are given as follows:
\begin{align}
& \frac{\partial \mathcal{L}}{\partial w}=0 \rightarrow w=\sum_{i=1}^n \alpha_i y_i \phi\left(x_i\right), \\
& \frac{\partial \mathcal{L}}{\partial b}=0 \rightarrow \sum_{i=1}^n \alpha_i y_i=0, \\
& \frac{\partial \mathcal{L}}{\partial \xi_i}=0 \rightarrow \alpha_i=C \mathcal{M}_i \xi_i, \quad i=1, \ldots, n, \\
& \frac{\partial \mathcal{L}}{\partial \alpha_i}=0 \rightarrow y_i\left[w^{\top} \phi\left(x_i\right)+b\right]-1+\xi_i=0, \quad i=1, \ldots, n.
\end{align}
After the elimination of the variables $w$ and $\xi$, a system of linear equations can be obtained as follows:
\begin{align}
\left[\begin{array}{cc}
0 & -Y^{\top} \\
Y & \Omega+(\mathcal{M} C)^{-1}
\end{array}\right]\left[\begin{array}{l}
b \\
\alpha
\end{array}\right]=\left[\begin{array}{l}
0 \\
1_n
\end{array}\right],
\end{align}
where $1_n=[1, 1, \ldots, 1] \in R^n$, $(\mathcal{M}C)^{-1}=\operatorname{diag}\left\{\frac{1}{\mathcal{M}_1 C}, \frac{1}{\mathcal{M}_2 C}, \ldots, \frac{1}{\mathcal{M}_n C}\right\}$, $\alpha=\left[\alpha_1, \alpha_2, \ldots, \alpha_n\right]$, $Y=\left[y_1, y_2, \ldots, y_n\right]$, and $\Omega_{i j} = y_i y_j \mathcal{K}\left(x_i, x_j\right)$, with $\mathcal{K}$ the kernel function. After obtaining the optimal $\alpha$ and $b$, the following decision function can be utilized to predict the label of a new sample $\hat{x}$.
\vspace{-0.3cm}
\begin{align} \label{decision function}
\hat{y}= \operatorname{sign}\left(\sum_{i=1}^n y_i \alpha_i \mathcal{K}\left(x_i, \hat{x}\right)+b\right).
\end{align}
\vspace{2mm}

\noindent
\textbf{Dual approach to class-center determination:}\\
The process of determining the class-center is pivotal in assigning the membership scheme within the proposed Flexi-Fuzz-LSSVM model. Traditional methods often compute the class-center by taking the arithmetic mean of all samples associated with that class. However, the mean may not be a robust estimator, particularly in the presence of outliers or non-symmetrical data distributions \cite{crow1967robust}. Outliers, being extreme values, can significantly skew the mean, resulting in an inaccurate representation of the true class-center. To address this limitation, the median is introduced as an alternative estimator. Unlike the mean, the median is a more robust measure of central tendency, as it is less affected by outliers and provides a more stable and reliable location estimator \cite{rousseeuw1991tutorial}. The stability of the median in the face of skewed data distributions makes it a superior choice for accurately determining the class-center in noisy environments, particularly in the context of AD diagnosis, where data variability and noise are prevalent challenges. In the proposed Flexi-Fuzz-LSSVM model, we employ both the conventional mean approach and the innovative median approach to determine the class-center.\\
\textbf{Key remarks:}\\
1) The adoption of two distinct strategies for initializing class-centers culminates in the development of two Flexi-Fuzz LSSVM model variants. Consequently, it enhances the diversity of our proposed approach.\\
2) The models utilizing the mean and median techniques for center determination are designated as Flexi-Fuzz LSSVM-I and Flexi-Fuzz LSSVM-II, respectively, each offering unique advantages tailored to different data conditions.

\section{Experimental Results} \label{Experimental results}
To assess the effectiveness of the proposed models, i.e., FlexiFuzz-LSSVM-I and FlexiFuzz-LSSVM-II, we evaluate their performance against baseline models including SVM \cite{cortes1995support}, LSSVM \cite{suykens1999least}, FSVM \cite{lin2002fuzzy}, FSVM-CIL-Lin \cite{batuwita2010fsvm}, FSVM-CIL-Exp \cite{batuwita2010fsvm}, ACFSVM \cite{tao2020affinity}, and IF-RVFL \cite{malik2022alzheimer}. We utilized publicly available UCI \cite{dua2017uci} and KEEL \cite{derrac2015keel} benchmark datasets. Moreover, we deploy the proposed models on the AD dataset, accessible through the Alzheimer's Disease Neuroimaging Initiative (ADNI) ($adni.loni.usc.edu$). The detailed experimental setup employed for evaluating the models is provided in Section S.II of the supplement file.

\subsection{Evaluation on UCI and KEEL Datasets}
In this subsection, we provide a comprehensive analysis and comparison of the proposed Flexi-Fuzz-LSSVM-I and Flexi-Fuzz-LSSVM-II models against various baseline models. This comparison spans $30$ benchmark datasets from the UCI and KEEL repositories, covering a diverse range of domains. The results, including average accuracy and rank, are summarized in Table \ref{UCI and KEEL Results without noise}, while detailed classification accuracies along with the optimal parameters for each dataset are presented in Table S.I of the supplementary file. The average accuracies of the existing baseline models including SVM, LSSVM, FSVM, FSVM-CIL-Lin, FSVM-CIL-Exp, ACFSVM, and IF-RVFL are $89\%$, $89.96\%$, $88.82\%$, $88.87\%$, $89.14\%$, $86.52\%$, and $86.11\%$ respectively. In contrast, the proposed Flexi-Fuzz-LSSVM-I and Flexi-Fuzz-LSSVM-II models achieved average accuracies of $90.02\%$ and $90.30\%$, respectively. This demonstrates a clear performance improvement, with the proposed Flexi-Fuzz-LSSVM-II model securing the top position and the Flexi-Fuzz-LSSVM-I model achieving the second position in terms of average accuracy.
Further, we can deduce that the innovative median approach used in Flexi-Fuzz-LSSVM-II to determine the class-center proves to be more effective in dealing with real-world data. The superiority of the median approach lies in its robustness to outliers. Unlike the mean, which can be significantly influenced by extreme values, the median provides a more stable central tendency measure, especially in skewed distributions. This characteristic is particularly advantageous in real-world datasets where noise and outliers are prevalent.

Relying solely on average accuracy as a single metric could be problematic, as exceptional performance on specific datasets might mask inadequate performance on others. To mitigate this concern, it becomes essential to individually rank each model with respect to each dataset, allowing for a comprehensive assessment of their respective capabilities. In the ranking scheme \cite{demvsar2006statistical}, the model with the poorest performance on a dataset is assigned a higher rank, while the model achieving the best performance is assigned a lower rank. In the assessment of $p$ models across $\mathcal{D}$ datasets, the rank of the $j^{th}$ model on the $i^{th}$ dataset can be represented as $\mathcal{R}_j^i$. Then the average rank of the model is determined as follows: $\mathscr{R}_j = \frac{1}{\mathcal{D}}\sum_{i=1}^{\mathcal{D}}\mathcal{R}_j^i$. The average rank of the proposed  Flexi-Fuzz-LSSVM-I and Flexi-Fuzz-LSSVM-II along with the baseline SVM, LSSVM, FSVM, FSVM-CIL-Lin, FSVM-CIL-Exp, ACFSVM, and IF-RVFL are $3.90$, $3.35$, $5.45$, $3.77$, $5.23$, $5.55$, $5.13$, $6.72$, and $5.90$ respectively. The proposed Flexi-Fuzz-LSSVM-II model achieved the lowest average rank among all the models, while the proposed Flexi-Fuzz-LSSVM-I ranked third. As a lower rank indicates a better-performing model, the proposed Flexi-Fuzz-LSSVM-II emerged as the best-performing model. We now proceed to conduct statistical tests to ascertain the significance of the results. Firstly, we utilize the Friedman test \cite{demvsar2006statistical} to ascertain if there exist noteworthy distinctions among the models. The null hypothesis posits that the models exhibit equal performance, as indicated by their average rank. The Friedman test conforms to the chi-squared distribution $\chi_F^2$ with $(p-1)$ degrees of freedom (d.o.f) and is expressed as: $\chi_F^2 = \frac{12\mathcal{D}}{p(p+1)}\left[\sum_j \mathscr{R}_j^2 - \frac{p(p+1)^2}{4}\right]$. However, the Friedman statistic is overly cautious in nature. To address this, \citet{iman1980approximations} introduced a more robust statistic: $F_F = \frac{(\mathcal{D}-1)\chi_F^2}{\mathcal{D}(p-1)-\chi_F^2}$, which follows $F$ distribution with $((p-1),(p-1)(\mathcal{D}-1))$ d.o.f.. For $p=9$ and $\mathcal{D} = 30$, we obtained $\chi_F^2 = 39.15$ and $F_F = 5.65$. From the $F$-distribution table at $5\%$ level of significance $F_F(8, 232) = 1.9784$.
Since, $F_F > 1.9784$, thus we reject
the null hypothesis. Hence, substantial differences exist among the models. Next, we employ the Nemenyi post hoc test \cite{demvsar2006statistical} to assess the pairwise differences among the models. The critical difference ($C.D.$) is determined as $C.D.=q_{\alpha}\sqrt{\frac{p(p+1)}{6\mathcal{D}}}$, where $q_{\alpha}$ denotes the critical value obtained from the distribution table for the two-tailed Nemenyi test. Referring to the statistical $F$-distribution table, where $q_\alpha =  3.102$ at a $5\%$ significance level, the $C.D.$ is computed as $2.1934$. The average rank differences between the proposed Flexi-Fuzz-LSSVM-I and Flexi-Fuzz-LSSVM-II models with the baseline SVM, LSSVM, FSVM, FSVM-CIL-Lin, FSVM-CIL-Exp, ACFSVM, and IF-RVFL models are $(1.55, 2.10)$, $(0.13, 0.42)$, $(1.33, 1.88)$, $(1.65, 2.20)$, $(1.23, 1.78)$, $(2.82, 3.37)$, and $(2, 2.55)$ respectively. As per the Nemenyi post hoc test, the proposed Flexi-Fuzz-LSSVM-II exhibits significant differences compared to the FSVM-CIL-Lin, ACFSVM, and IF-RVFL models. The proposed Flexi-Fuzz-LSSVM-I does not exhibit a statistical difference with the baseline models except ACFSVM. However, the proposed Flexi-Fuzz-LSSVM-I model surpasses the baseline models in terms of the average rank. Taking into account all these findings, we can conclude that the proposed Flexi-Fuzz-LSSVM-I demonstrates a competitive nature, and the proposed Flexi-Fuzz-LSSVM-II showcases superior performance against the existing models. The evaluation on the UCI and KEEL datasets with the added label noise is discussed thoroughly in Section S.II.B of the supplement file.

\begin{table*}[htp]
\centering
    \caption{Comparison of average results between the proposed Flexi-Fuzz-LSSVM models and baseline models on UCI and KEEL datasets without label noise.}
    \label{UCI and KEEL Results without noise}
     {\resizebox{1.00\linewidth}{!}{
\begin{tabular}{lccccccccc}
\hline
Dataset & SVM \cite{cortes1995support} & LSSVM \cite{suykens1999least} & FSVM \cite{lin2002fuzzy} & FSVM-CIL-Lin \cite{batuwita2010fsvm} & FSVM-CIL-Exp \cite{batuwita2010fsvm} & ACFSVM \cite{tao2020affinity}  &  IF-RVFL \cite{malik2022alzheimer} & Flexi-Fuzz-LSSVM-I$^{\dagger}$ & Flexi-Fuzz-LSSVM-II$^{\dagger}$ \\ \hline
Average ACC & $89$ & $89.96$ & $88.82$ & $88.87$ & $89.14$ & $86.52$  &  $86.11$ & $\underline{90.02}$ & $\textbf{90.30}$ \\ \hline
Average Rank & $5.45$ & $\underline {3.77}$ & $5.23$ & $5.55$ & $5.13$ & $6.72$ & $5.90$ & $3.90$ & $\textbf{3.35}$ \\ \hline
\multicolumn{9}{l}{$^{\dagger}$ represents the proposed models.}\\ 
\multicolumn{9}{l}{Bold and underlined text denote the models with the highest and second-highest average accuracy, respectively.}
\end{tabular}}}
\end{table*}

\begin{table*}[htp]
\centering
\caption{Performance comparison between the proposed Flexi-Fuzz-LSSVM models and baseline models on the ADNI dataset.}
\label{AD Results}
{ \resizebox{1.0\linewidth}{!}{
\begin{tabular}{cccccccccc}
\hline
{Subjects} & SVM \cite{cortes1995support} & LSSVM \cite{suykens1999least} & FSVM \cite{lin2002fuzzy} & FSVM-CIL-Lin \cite{batuwita2010fsvm} & FSVM-CIL-Exp \cite{batuwita2010fsvm} & ACFSVM \cite{tao2020affinity}  &  IF-RVFL \cite{malik2022alzheimer} & Flexi-Fuzz-LSSVM-I$^{\dagger}$ & Flexi-Fuzz-LSSVM-II$^{\dagger}$ \\
 (no. of samples $\times$ no. of features)& (ACC, Sensitivity) & (ACC, Sensitivity) & (ACC, Sensitivity) & (ACC, Sensitivity) & (ACC, Sensitivity) & (ACC, Sensitivity) & (ACC, Sensitivity) & (ACC, Sensitivity) & (ACC, Sensitivity) \\
 & (Specificity, Precision) & (Specificity, Precision)  & (Specificity, Precision)  & (Specificity, Precision) & (Specificity, Precision) & (Specificity, Precision) & (Specificity, Precision) & (Specificity, Precision) & (Specificity, Precision) \\ \hline
CN vs AD & $(79.03, 55.56)$ & $(84.68, 88.89)$ & $(79.03, 55.56)$ & $(79.03, 55.56)$ & $(80.65, 68.52)$& $(62.1, 57.41)$ & $(82.27, 82.36)$ & $(86.29, 72.22)$ & $(87.9, 88.89)$ \\
$(415 \times 91)$ & $(97.14, 93.75)$ & $(88.57, 85.71)$ & $(97.14, 93.75)$ & $(97.14, 93.75)$ & $(90, 84.09)$ & $(65.71, 56.36)$ & $(82.46, 65.35)$ & $(97.14, 95.12)$ & $(87.14, 84.21)$ \\
CN vs MCI & $(61.5, 54.39)$ & $(61.57, 58.77)$ & $(62.5, 54.39)$ & $(61.5, 54.39)$ & $(61.5, 54.39)$& $(45.45, 46.49)$ & $(67.81, 72.36)$  & $(67.91, 85.09)$ & $(68.98, 87.72)$ \\
$(626 \times 91)$ & $(72.6, 75.61)$ & $(68.49, 74.44)$ & $(72.6, 75.61)$ & $(72.6, 75.61)$ & $(72.6, 75.61)$ & $(43.84, 56.38)$  & $(65.25, 68.37)$ & $(41.1, 69.29)$ & $(39.73, 69.44)$ \\
MCI vs AD & $(69.71, 42.86)$ & $(71.43, 77.5)$ & $(69.71, 42.86)$ & $(71.71, 42.86)$ & $(69.57, 54.64)$ & $(68, 58.36)$ & $(70.33, 56.38)$ & $(68.57, 81.51)$ & $(71.71, 37.5)$ \\
$(585 \times 91)$ & $(88.24, 63.16)$ & $(87.39, 58.33)$ & $(88.24, 63.16)$ & $(88.24, 63.16)$ & $(85.71, 59.52)$ & $(80, 68.25)$ & $(72.36, 64)$ & $(51.11, 41.07)$ & $(84.87, 53.85)$ \\ \hline
Average ACC & $70.08$ & $72.56$ & $70.41$ & $70.75$ & $70.57$ &  $58.52$ & $73.47$  & $\underline{74.26}$ & $\textbf{76.20}$ \\ \hline
\multicolumn{9}{l}{$^{\dagger}$ represents the proposed models.}\\
\multicolumn{9}{l}{Bold and underlined text denote the models with the highest and second-highest average accuracy, respectively.}
\end{tabular}}}
\end{table*}

\subsection{Evaluation on ADNI dataset}
AD is a progressive neurological disorder that detrimentally affects memory and cognitive functions. AD commonly begins with mild cognitive impairment (MCI). Currently, the exact causes of AD remain incompletely comprehended. However, the precise identification and diagnosis of AD are crucial in the provision of patient treatment, particularly during the first phase. In this study, we utilized scans from the Alzheimer’s Disease Neuroimaging Initiative (ADNI) dataset to train the proposed Flexi-Fuzz-LSSVM-I and Flexi-Fuzz-LSSVM-II models.
The ADNI project, launched by Michael W. Weiner in 2003, seeks to evaluate a range of neuroimaging techniques, such as positron emission tomography (PET), magnetic resonance imaging (MRI), and other diagnostic assessments for AD, particularly during the MCI phase. The feature extraction pipeline employed in this study aligns with the methodology described in \cite{richhariya2021efficient}. The dataset encompasses three classification scenarios: control normal (CN) versus AD, MCI versus AD, and CN versus MCI.
\par
The performance metrics of the proposed Flexi-Fuzz-LSSVM-I and Flexi-Fuzz-LSSVM-II models, in comparison to baseline models, are presented in Table \ref{AD Results}. Notably, Flexi-Fuzz-LSSVM-II and Flexi-Fuzz-LSSVM-I secured the top first and second positions with average accuracies of $76.20\%$ and $74.26\%$, respectively. In contrast, the baseline models, including SVM, LSSVM, FSVM, FSVM-CIL-Lin, and FSVM-CIL-Exp, demonstrated lower average accuracies of $70.08\%$, $72.56\%$, $70.41\%$, $70.75\%$, and $70.57\%$, respectively. Compared to the third-top model, LSSVM, the proposed models (Flexi-Fuzz-LSSVM-I and Flexi-Fuzz-LSSVM-II) surpass it by approximately $1.7\%$ and $3.64\%$ in average accuracy, respectively. The Flexi-Fuzz-LSSVM-II model achieves the highest accuracy of $87.9\%$ for the CN vs AD case, followed by Flexi-Fuzz-LSSVM-I with an accuracy of $86.29\%$. For the CN vs MCI case, Flexi-Fuzz-LSSVM-II and Flexi-Fuzz-LSSVM-I again emerge as the top performers, with average accuracies of $68.98\%$ and $67.91\%$, respectively. In the MCI vs AD case, Flexi-Fuzz-LSSVM-II and the existing FSVM-CIL-Lin stand out as the most accurate classifiers, with an accuracy of $71.71\%$. Thus, Flexi-Fuzz-LSSVM-II consistently demonstrates superior performance, achieving high accuracy across various cases and establishing its prominence among the models. The success of the Flexi-Fuzz-LSSVM-II model can be largely attributed to its innovative use of the median approach to determine the class-center. Unlike the traditional mean approach, the median provides a more stable reference point that is less influenced by outliers and noise. This stability greatly enhances the model's robustness, enabling it to perform well even under real-world conditions. This advancement is particularly beneficial in the context of AD diagnosis, where variability and noise in data are common challenges. Moreover, to illustrate the comparison between the proposed Flexi-Fuzz-LSSVM models and the baseline models in terms of specificity, sensitivity, and precision, we plotted the bar graphs shown in Figure S.4 of the supplement file. The overall findings underscore the effectiveness of the proposed models in distinguishing between different cognitive states.

\section{Conclusions and Future Work} \label{Conclusion}
In this paper, we introduced Flexi-Fuzz, a robust and flexible membership scheme designed to address the challenges of noise, outliers, and class imbalance commonly encountered in AD diagnosis. By integrating the Flexi-Fuzz scheme into the framework of least squares support vector machines (LSSVM) and employing both the traditional mean approach and the innovative median approach to determine the class-center, we developed two novel models: Flexi-Fuzz-LSSVM-I and Flexi-Fuzz-LSSVM-II. The success of the proposed models is due to their robust membership scheme, which distinguishes noisy samples from normal ones while retaining the influence of boundary samples, the adaptive adjustments enabled by the flexible parameter \(\lambda\) and \(k\)-nearest neighborhood parameter \(k\), and the median approach in Flexi-Fuzz-LSSVM-II, which enhances robustness by providing a stable reference less affected by outliers and noise. Extensive experimental evaluation reflects that the proposed models significantly outperform baseline models on UCI and KEEL datasets, maintaining exceptional performance even with introduced label noise datasets. Additionally, the efficacy of our models in diagnosing Alzheimer's disease (AD) was validated using the ADNI dataset. Flexi-Fuzz-LSSVM-I exhibited competitive performance, while Flexi-Fuzz-LSSVM-II achieved the highest average accuracy and classification accuracy across all three cases (CN vs AD, CN vs MCI, and MCI vs AD). However, one limitation of the proposed Flexi-Fuzz-LSSVM models lies in the need for manual tuning of the two hyperparameters, \(\lambda\) and \(k\), which are integral to the membership scheme. These hyperparameters require careful adjustment to optimize the performance for different datasets and noise levels, which can be time-consuming.
\par
In future work, researchers could focus on developing adaptive methods to dynamically and efficiently adjust \(\lambda\) and \(k\) during the training process, thereby eliminating the need for manual tuning. Such methods could leverage meta-heuristic optimization algorithms, or self-adaptive mechanisms to automate the adjustment process. Further, one can leverage the Flexi-Fuzz membership scheme and amalgamate it with cutting-edge models to tackle complex real-world problems. 

\section*{Acknowledgment}
This work is supported by Science and Engineering Research Board (SERB) through the MTR/2021/000787 grant as part of the Mathematical Research Impact-Centric Support (MATRICS) scheme. Mushir Akhtar’s research was supported by a fellowship grant (no. 09/1022(13849)/2022-EMR-I) from the Council of Scientific and Industrial Research (CSIR), New Delhi. 
Data collection and sharing for this project were supported by funding from the Alzheimer's Disease Neuroimaging Initiative (ADNI), provided by the National Institutes of Health Grant U01 AG024904, and the Department of Defense award number W81XWH-12-2-0012. ADNI is financed by the National Institute on Aging, the National Institute of Biomedical Imaging and Bioengineering, and generous contributions from various organizations, including AbbVie, the Alzheimer’s Association, the Alzheimer’s Drug Discovery Foundation, Araclon Biotech, BioClinica, Inc., Biogen, Bristol-Myers Squibb Company, CereSpir, Inc., Cogstate, Eisai Inc., Elan Pharmaceuticals, Inc., Eli Lilly and Company, EuroImmun, F. Hoffmann-La Roche Ltd and its affiliated company Genentech, Inc., Fujirebio, GE Healthcare, IXICO Ltd., Janssen Alzheimer Immunotherapy Research \& Development, LLC., Johnson \& Johnson Pharmaceutical Research \& Development LLC., Lumosity, Lundbeck, Merck \& Co., Inc., Meso Scale Diagnostics, LLC., NeuroRx Research, Neurotrack Technologies, Novartis Pharmaceuticals Corporation, Pfizer Inc., Piramal Imaging, Servier, Takeda Pharmaceutical Company, and Transition Therapeutics. The Canadian Institutes of Health Research funds ADNI clinical sites in Canada. Private sector donations are managed by the Foundation for the National Institutes of Health (www.fnih.org). The grantee organization is the Northern California Institute for Research and Education, with the study coordinated by the Alzheimer’s Therapeutic Research Institute at the University of Southern California. ADNI data are distributed by the Laboratory for Neuro Imaging at the University of Southern California.

\bibliographystyle{IEEEtranN}
\bibliography{refs.bib}

\end{document}


\title{Supplementary Material for the Manuscript ``Flexi-Fuzz least squares SVM for Alzheimer's diagnosis: Tackling noise, outliers, and class imbalance"}
\author{Mushir Akhtar, \IEEEmembership{Graduate Student Member,~IEEE},  A. Quadir, \IEEEmembership{Graduate Student Member,~IEEE}, M. Tanveer{$^*$}, \IEEEmembership{Senior Member,~IEEE}, Mohd. Arshad, for the Alzheimer’s Disease Neuroimaging Initiative{$^{**}$}
\thanks{ \noindent $^*$Corresponding Author\\
    Mushir Akhtar,  A. Quadir, M. Tanveer and Mohd. Arshad are with the Department of Mathematics, Indian Institute of Technology Indore, Simrol, Indore, 453552, India (e-mail: phd2101241004@iiti.ac.in, mscphd2207141002@iiti.ac.in, mtanveer@iiti.ac.in, arshad@iiti.ac.in).\\  $^{**}$Data utilized in this article were sourced from the Alzheimer’s Disease Neuroimaging Initiative (ADNI) database (adni.loni.usc.edu). While the investigators within ADNI contributed to the design and implementation of ADNI and/or provided data, they were not involved in the analysis or writing of this paper.
 }}

\maketitle

\section{Related Work}
In this section, we briefly discuss the formulation of support vector machine (SVM) \cite{cortes1995support}, least squares support vector machine (LSSVM) \cite{suykens1999least}, and fuzzy support vector machine (FSVM) \cite{lin2002fuzzy}.
\subsection{Support Vector Machine} 
The classical Support Vector Machine \cite{cortes1995support} classifier (SVM) emerges from the geometrical idea of maximizing the margin between the separating hyperplane and both classes in binary classification problems. Formally, it is defined as the solution of the following optimization problem:
\begin{align}
 \min _{w, b} \hspace{0.2cm} &\frac{1}{2}\|w\|_2^2+C \sum_{i=1}^n \xi_i \nonumber\\
 \text { s.t. } &y_i\left(w^{\top} \phi\left(x_i\right)+b\right) \geq 1-\xi_i, \quad i=1, \ldots, n, \nonumber\\
&\xi_i \geq 0, \quad i=1, \ldots, n,
\end{align}
where $C>0$ is the regularization parameter and $w \in \mathbb{R}^p$ is the weight vector that defines the separating hyperplane. Through standard Lagrangian duality, the corresponding dual optimization problem turns out to be:
\begin{align}
 \min _{\boldsymbol{\alpha}}\hspace{0.2cm} &\frac{1}{2} \boldsymbol{\alpha}^{\top} \tilde{\mathbf{K}} \boldsymbol{\alpha}-\mathbf{1}^{\top} \boldsymbol{\alpha} \nonumber\\
\text { s.t. } &\sum_{i=1}^n y_i \alpha_i=0, \nonumber\\
& 0 \leq \alpha_i \leq C, \quad i=1, \ldots, n,
\end{align}
where $\alpha_i$'s are the Lagrange multipliers and $\tilde{\mathbf{K}}\in\mathbb{R}^{n \times n}$ corresponds to the labelled kernel matrix, $\tilde{k}_{i, j}=y_i y_j \mathcal{K}\left(x_i, x_j\right)$, with $\mathcal{K}$ the kernel function defined as $\mathcal{K}\left(x_i, x_j\right)=\phi\left(x_i\right) \cdot \phi\left(x_j\right)$. Once the dual problem has been solved and the bias $b$ has been deduced from the optimality conditions, the prediction for a new sample $x$ is given by:
$$
\operatorname{sign}\left(\sum_{i=1}^n y_i \alpha_i \mathcal{K}\left(x_i, x\right)+b\right)
$$

\subsection{Least Squares Support Vector Machine}
LSSVM \cite{suykens1999least} is a least-squares version of the SVM classifier. The main idea underlying LSSVM involves transforming SVM's original inequality constraints into equality constraints, thereby mitigating the computational complexity associated with solving a quadratic programming problem. The strategic adaptation of SVM constraints into an equivalent least-squares form in LSSVM not only simplifies the computational burden but also offers a more efficient and scalable solution. The mathematical formulation of LSSVM is articulated as follows:
\begin{align} \label{LSSVM}
\min _{w, b}\hspace{0.2cm} &\frac{1}{2}\|w\|^2+\frac{1}{2} C \sum_{i=1}^n \xi_i^2 \nonumber\\
\text {s.t.} \hspace{0.2cm} &y_i\left(w^{\top} \phi\left(x_i\right)+b\right)=1-\xi_i, \quad i=1,2, \ldots,n.
\end{align}
The Lagrangian for equation (\ref{LSSVM}) can be defined as:
\begin{align} \label{lagrange_LSSVM}
\mathcal{L}(w, b, \xi ; \alpha)=&\frac{1}{2}\|w\|^2+\frac{1}{2} C \sum_{i=1}^n \xi_i^2 \nonumber \\
&-\sum_{i=1}^n \alpha_i\left\{y_i\left[w^{\top} \phi\left(x_i\right)+b\right]-1+\xi_i\right\},
\end{align}
where $\alpha_i$'s denote the Lagrange multipliers. The solution of equation (\ref{lagrange_LSSVM}) can be derived by taking partial derivatives of $\mathcal{L}$ with respect to $w, b, \xi_i, \alpha_i$, and subsequently equating the resulting expressions to zero. For detailed derivation, refer to \cite{suykens1999least}.

\begin{figure*}[ht!]
\begin{minipage}{.45\linewidth}
\centering
\subfloat[ecoli-0-1-4-6\_vs\_5]{\includegraphics[scale=0.42]{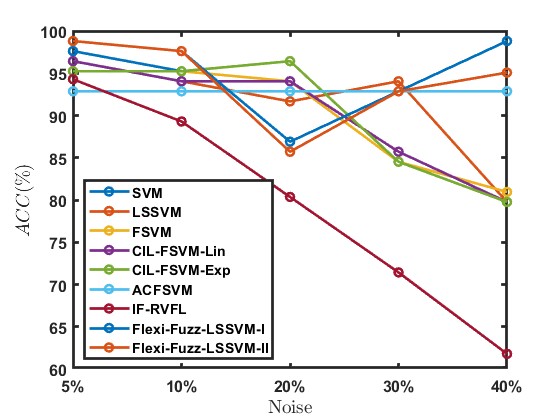}}
\end{minipage}
\begin{minipage}{0.45\linewidth}
\centering
\subfloat[ecoli-0-3-4-7\_vs\_5-6]{\includegraphics[scale=0.42]{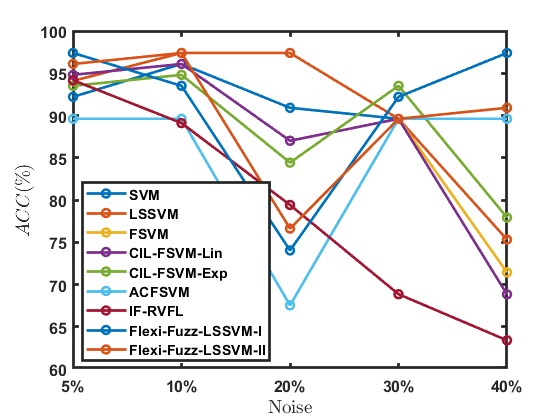}}
\end{minipage}
\par\medskip
\begin{minipage}{.45\linewidth}
\centering
\subfloat[votes]{\includegraphics[scale=0.42]{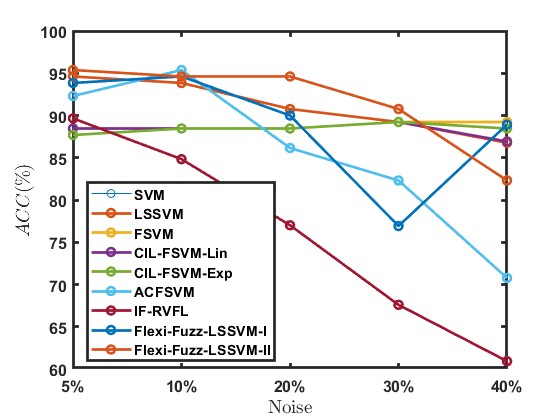}}
\end{minipage}
\begin{minipage}{.45\linewidth}
\centering
\subfloat[yeast3]{\includegraphics[scale=0.42]{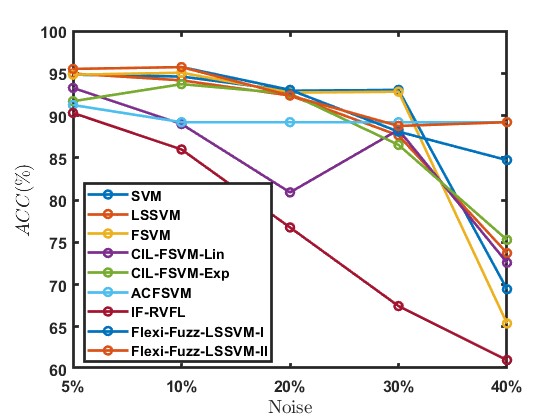}}
\end{minipage}
\caption{Illustrates the impact of various levels of label noise on the performance of the proposed Flexi-Fuzz-LSSVM models compared to the baseline models, demonstrating their robustness against label noise. The datasets used are (a) ecoli-0-1-4-6\_vs\_5, (b) ecoli-0-3-4-7\_vs\_5-6, (c) votes, and (d) yeast3.}
\label{Effect of different labels of noise on the performance of the proposed Flexi-Fuzz-LSSVM-I and Flexi-Fuzz-LSSVM-II model}
\end{figure*}

\subsection{Fuzzy Support Vector Machine}
In contrast to conventional SVM, fuzzy SVM (FSVM) \cite{lin2002fuzzy} incorporates a distinctive feature in its objective function by considering the belongingness of samples to a class in a fuzzy manner. The objective function for FSVM can be defined as follows:
\begin{align}\label{FuzzySVM}
 \min _{w, b} \hspace{0.2cm} &\frac{1}{2}\|w\|^2+C \sum_{i=1}^n m_i \xi_i \nonumber\\
 \text { s.t. } &y_i\left(w^{\top} \phi\left(x_i\right)+b\right) \geq 1-\xi_i, \quad i=1, \ldots, n, \nonumber \\
&\xi_i \geq 0, \hspace{0.5cm}  0 \leq m_i \leq 1, \quad i=1, \ldots, n,
\end{align}
where $m_i$ is the membership value of $i^{th}$ sample to describe belongingness to its own class. The dual form of equation (\ref{FuzzySVM}) can be described as follows:
\begin{align}
 \min _{\boldsymbol{\alpha}}\hspace{0.2cm} &\frac{1}{2} \boldsymbol{\alpha}^{\top} \tilde{\mathbf{K}} \boldsymbol{\alpha}-\mathbf{1}^{\top} \boldsymbol{\alpha} \nonumber\\
\text { s.t. } &\sum_{i=1}^n y_i \alpha_i=0, \nonumber\\
& 0 \leq \alpha_i \leq m_i C, \quad i=1, \ldots, n.
\end{align}
Here, $\alpha_i$'s and $\tilde{\mathbf{K}}$ hold the same significance as previously defined in the context of SVM.

It can be deduced from (\ref{FuzzySVM}) that the relevance of a training sample's membership value manifests only when the sample is misclassified. As previously discussed, a drawback of SVM lies in its limited effectiveness in handling outliers. In contrast, FSVM, when equipped with well-defined membership functions, exhibits enhanced efficacy in addressing outliers. Unlike SVM, FSVM acknowledges and accommodates diverse sample contributions to the classifier formation, as delineated in the definition of its objective function. This distinctive feature empowers FSVM to adeptly handle outliers through the utilization of precisely defined membership functions. Consequently, the development of suitable heuristics for membership functions emerges as a pivotal factor for enhancing the generalization performance of SVM.

\begin{figure*}[ht!]
\begin{minipage}{0.45\linewidth}
\centering
\subfloat[brwisconsin (Flexi-Fuzz-LSSVM-I)]{\includegraphics[scale=0.42]{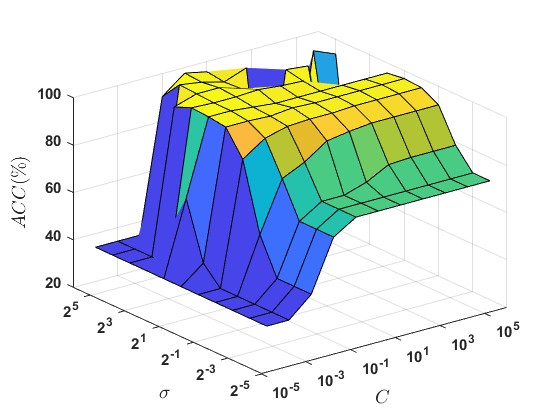}}
\end{minipage}
\begin{minipage}{0.45\linewidth}
\centering
\subfloat[breast\_cancer\_wisc (Flexi-Fuzz-LSSVM-I)]{\includegraphics[scale=0.42]{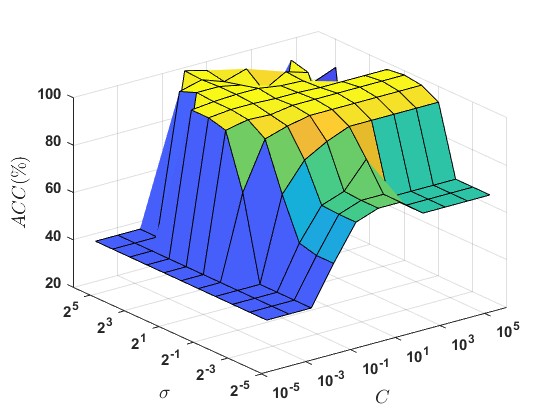}}
\end{minipage}
\par\medskip
\begin{minipage}{0.45\linewidth}
\centering
\subfloat[brwisconsin (Flexi-Fuzz-LSSVM-II)]{\includegraphics[scale=0.42]{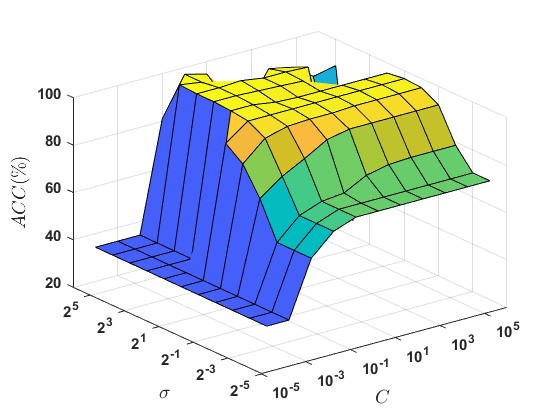}}
\end{minipage}
\begin{minipage}{0.45\linewidth}
\centering
\subfloat[breast\_cancer\_wisc (Flexi-Fuzz-LSSVM-II)]{\includegraphics[scale=0.42]{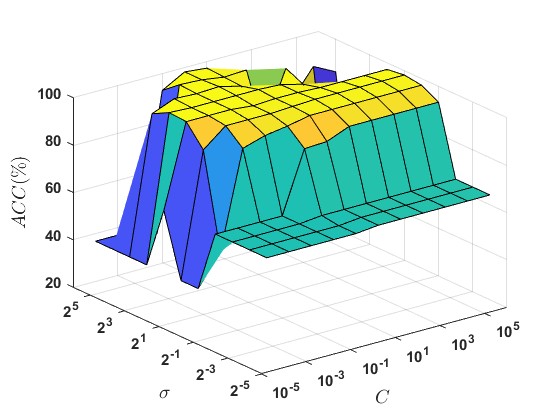}}
\end{minipage}
\caption{Demonstrates the effect of the regularization parameter \(C\) and the kernel parameter \(\sigma\) on the performance of the proposed Flexi-Fuzz-LSSVM models. The datasets used for this analysis are (a) brwisconsin with Flexi-Fuzz-LSSVM-I, (b) breast\_cancer\_wisc with Flexi-Fuzz-LSSVM-I, (c) brwisconsin with Flexi-Fuzz-LSSVM-II, and (d) breast\_cancer\_wisc with Flexi-Fuzz-LSSVM-II. The surface plots illustrate how changes in \(C\) and \(\sigma\) influence the accuracy of the models, highlighting the sensitivity of the Flexi-Fuzz-LSSVM models to these parameters.}
\label{effect of C and sigma parameter}
\end{figure*}
\begin{figure*}[ht!]
\begin{minipage}{.45\linewidth}
\centering
\subfloat[ecoli-0-6-7\_vs\_3-5 (Flexi-Fuzz-LSSVM-I)]{\includegraphics[scale=0.42]{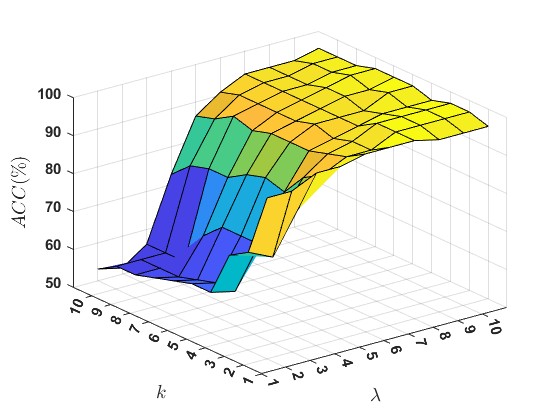}}
\end{minipage}
\begin{minipage}{.45\linewidth}
\centering
\subfloat[statlog-german-credit (Flexi-Fuzzz-LSSVM-I)]{\includegraphics[scale=0.42]{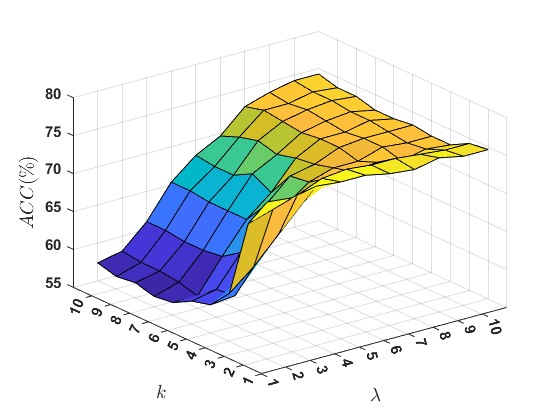}}
\end{minipage}
\par\medskip
\begin{minipage}{.45\linewidth}
\centering
\subfloat[ecoli-0-6-7\_vs\_3-5 (Flexi-Fuzz-LSSVM-II)]{\includegraphics[scale=0.42]{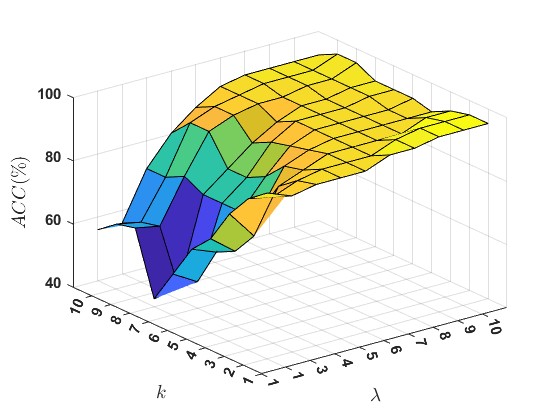}}
\end{minipage}
\begin{minipage}{.45\linewidth}
\centering
\subfloat[statlog-german-credit (Flexi-Fuzz-LSSVM-II)]{\includegraphics[scale=0.42]{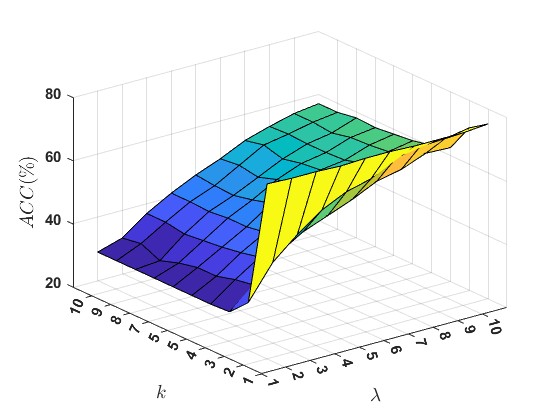}}
\end{minipage}
\caption{Explores the impact of the flexible parameter \(\lambda\) and the \(k\)-nearest neighborhood parameter \(k\) on the performance of the proposed Flexi-Fuzz-LSSVM models. The datasets analyzed are (a) ecoli-0-6-7\_vs\_3-5 with Flexi-Fuzz-LSSVM-I, (b) statlog-german-credit with Flexi-Fuzz-LSSVM-I, (c) ecoli-0-6-7\_vs\_3-5 with Flexi-Fuzz-LSSVM-II, and (d) statlog-german-credit with Flexi-Fuzz-LSSVM-II. The surface plots illustrate how variations in \(\lambda\) and \(k\) parameters affect the models' accuracy, highlighting the sensitivity and influence of these parameters on the performance of the Flexi-Fuzz-LSSVM models.}
\label{effect of lambda and k parameter}
\end{figure*}

\section{Experimental Results}
In this section, we provide the experimental setup followed to evaluate the proposed Flexi-Fuzz-LSSVM models and baseline models. Further, we provide the evaluation on the UCI and KEEL datasets with added label noise.

\begin{figure*}[ht!]
\begin{minipage}{.340\linewidth}
\centering
\subfloat[CN vs AD]{\includegraphics[scale=0.33]{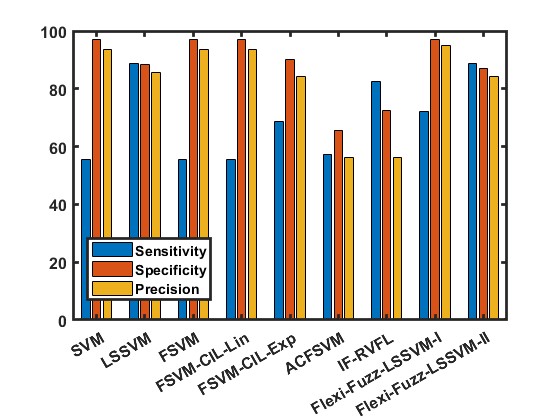}}
\end{minipage}
\begin{minipage}{.340\linewidth}
\centering
\subfloat[CN vs MCI]{\includegraphics[scale=0.33]{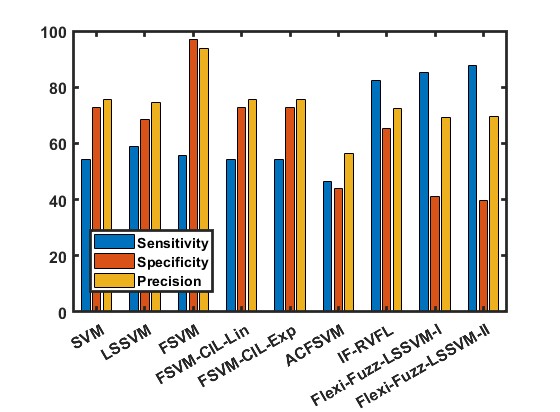}}
\end{minipage}
\begin{minipage}{.340\linewidth}
\centering
\subfloat[MCI vs AD]{\includegraphics[scale=0.33]{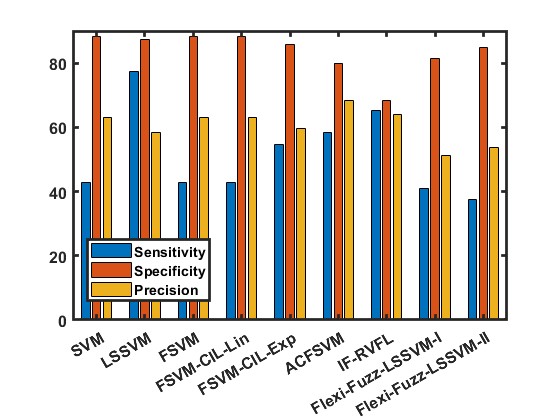}}
\end{minipage}
\caption{Compares the performance of the proposed Flexi-Fuzz-LSSVM models against baseline models using sensitivity, specificity, and precision metrics on the ADNI dataset. The comparisons are made for the following groups: (a) CN vs AD, (b) CN vs MCI, and (c) MCI vs AD.}
\label{Bar graph of specificity, sensitivity, and precision on ADNI dataset.}
\end{figure*}

\begin{table*}[ht!]
\centering
    \caption{Performance comparison of the proposed Flexi-Fuzz-LSSVM Models against the baseline models on UCI and KEEL datasets.}
    \label{UCI and KEEL Results without noise}
    \resizebox{1.00\linewidth}{!}{
\begin{tabular}{lccccccccc}
\hline
Dataset & SVM \cite{cortes1995support} & LSSVM \cite{suykens1999least} & FSVM \cite{lin2002fuzzy} & FSVM-CIL-Lin \cite{batuwita2010fsvm} & FSVM-CIL-Exp \cite{batuwita2010fsvm} & ACFSVM \cite{tao2020affinity} & IF-RVFL \cite{malik2022alzheimer} & Flexi-Fuzz-LSSVM-I$^{\dagger}$ & Flexi-Fuzz-LSSVM-II$^{\dagger}$ \\
 & ACC & ACC & ACC & ACC & ACC & ACC & ACC & ACC \\
 & $(C, \sigma)$ & $(C, \sigma)$ & $(C, \sigma)$ & $(C, \sigma)$ & $(C, \beta, \sigma)$ & $(C, \sigma)$  &  $(C, N, \sigma)$ & $(C, \lambda, k, \sigma)$ & $(C, \lambda, k, \sigma)$ \\ \hline
breast\_cancer\_wisc & $96.17$ & $98.09$ & $97.61$ & $96.65$ & $96.65$ & $98.09$ & $88.56$  & $98.56$ & $98.56$ \\
 & $(0.1, 0.5)$ & $(0.1, 8)$ & $(1, 1)$ & $(0.1, 1)$ & $(0.1, 0.4, 1)$ & $(1, 0.5)$ &  $(10000, 203, 2)$ & $(0.001, 1, 3, 8)$ & $(0.1, 2, 1, 8)$ \\
brwisconsin & $95.1$ & $97.06$ & $97.06$ & $95.59$ & $95.59$ & $95.59$ &  $90.64$ & $98.04$ & $98.04$ \\
 & $(0.1, 0.5)$ & $(0.1, 8)$ & $(1, 1)$ & $(10, 1)$ & $(100, 1, 1)$ & $(0.01, 1)$ &  $10000, 203, 2$ & $(0.01, 2, 1, 8)$ & $(0.01, 1, 4, 8)$ \\
chess\_krvkp & $92.59$ & $94.15$ & $92.59$ & $92.59$ & $93.32$ & $97.29$  &  $72.72$ & $96.97$ & $97.29$ \\
 & $(1, 0.5)$ & $(100000, 4)$ & $(10, 0.5)$ & $(10, 0.5)$ & $(1, 0.1, 0.5)$ & $(10, 0.5)$  &  $(100000, 203, 0.03125)$ & $(1, 7, 3, 16)$ & $(1, 10, 5, 16)$ \\
cleve & $80.9$ & $83.15$ & $80.9$ & $79.78$ & $80.9$ & $78.65$ &  $81.14$ &  $82.02$ & $80.9$ \\
 & $(1, 0.125)$ & $(1, 2)$ & $(10, 0.0625)$ & $(1, 0.25)$ & $(10, 1, 0.25)$ & $(10, 0.0625)$  & $(100000, 203, 32)$ & $(0.01, 2, 2, 4)$ & $(1, 1, 1, 8)$ \\
credit\_approval & $87.92$ & $88.41$ & $87.92$ & $87.92$ & $87.92$ & $85.99$ &  $86.23$ & $87.92$ & $88.41$ \\
 & $(10, 0.5)$ & $(100, 2)$ & $(10, 0.5)$ & $(10, 0.5)$ & $(10, 0.2, 0.5)$ & $(10, 0.5)$  &  $(100000, 183, 32)$ & $(0.1, 3, 2, 2)$ & $(0.01, 1, 1, 2)$ \\
echocardiogram & $87.18$ & $79.49$ & $89.74$ & $89.74$ & $87.18$ & $76.92$ &  $80.83$ & $82.05$ & $84.62$ \\
 & $(10, 0.25)$ & $(1, 16)$ & $(1, 0.25)$ & $(10, 0.5)$ & $(10, 0.9, 0.25)$ & $(0.1, 0.25)$ &   $(10000, 163, 8)$  & $(0.00001, 1, 1, 2)$ & $(0.1, 3, 1, 2)$ \\
ecoli-0-1-4-6\_vs\_5 & $96.43$ & $98.81$ & $97.62$ & $97.62$ & $97.62$ & $94.05$ & $98.93$ & $98.81$ & $98.81$ \\
 & $(1, 1)$ & $(10, 1)$ & $(10, 1)$ & $(100, 1)$ & $(100, 0.1, 1)$ & $(0.01, 1)$ &  $(1000, 183, 32)$ & $(0.0001, 1, 1, 1)$ & $(0.00001, 1, 1, 1)$ \\
ecoli-0-3-4-6\_vs\_5 & $96.72$ & $98.36$ & $98.36$ & $96.72$ & $96.72$ & $96.72$  &  $98.05$ & $98.36$ & $98.36$ \\
 & $(10, 0.125)$ & $(1, 2)$ & $(1000, 0.125)$ & $(10000, 0.125)$ & $(100, 0.1, 0.125)$ & $(0.01, 0.125)$ &  $(10000, 163, 32)$ & $(1, 1, 1, 4)$ & $(1, 1, 1, 4)$ \\
ecoli-0-3-4-7\_vs\_5-6 & $96.1$ & $97.4$ & $96.1$ & $94.81$ & $93.51$ & $92.21$ &  $97.67$ & $96.1$ & $94.81$ \\
 & $(10, 0.25)$ & $(10, 2)$ & $(100, 0.25)$ & $(1000, 0.25)$ & $(100, 0.6, 0.25)$ & $(0.01, 0.5)$  &  $(100000, 183, 2)$ & $(0.1, 9, 1, 4)$ & $(1, 6, 1, 4)$ \\
ecoli-0-6-7\_vs\_3-5 & $95.45$ & $96.97$ & $95.45$ & $93.94$ & $93.94$ & $92.42$  &  $96.85$ & $96.97$ & $96.97$ \\
 & $(1, 0.5)$ & $(10, 8)$ & $(1, 1)$ & $(10, 0.5)$ & $(10, 0.1, 0.5)$ & $(0.01, 0.5)$ &  $(100000, 123, 8)$ & $(0.1, 3, 6, 4)$ & $(0.01, 7, 2, 2)$ \\
ecoli-0-6-7\_vs\_5 & $95.45$ & $98.48$ & $92.42$ & $92.42$ & $93.94$ & $89.39$  &  $98.18$ & $98.48$ & $98.48$ \\
 & $(1, 1)$ & $(10, 4)$ & $(10, 1)$ & $(100, 1)$ & $(100, 0.9, 1)$ & $(0.01, 1)$  &  $(100, 183, 32)$ & $(1, 3, 1, 4)$ & $(1, 4, 1, 4)$ \\
ecoli3 & $94$ & $91$ & $90$ & $94$ & $96$ & $90$ &  $94.64$  & $92$ & $97$ \\
 & $(1, 1)$ & $(10, 0.5)$ & $(10, 1)$ & $(1000, 0.0625)$ & $(1000, 0.6, 0.0625)$ & $(0.00001, 1)$  &  $(100000, 143, 32)$ & $(10, 2, 6, 2)$ & $(10, 4, 6, 2)$ \\
hepatitis & $86.96$ & $89.13$ & $86.96$ & $86.96$ & $89.13$ & $80.43$  &  $85.81$ & $84.78$ & $89.13$ \\
 & $(10, 0.25)$ & $(10, 4)$ & $(1000, 0.25)$ & $(10000, 0.25)$ & $(10, 0.7, 0.25)$ & $(0.1, 0.5)$  &  $(100, 103, 2)$ & $(0.1, 2, 3, 2)$ & $(1, 4, 1, 8)$ \\
horse\_colic & $82.73$ & $85.45$ & $78.18$ & $79.09$ & $76.36$ & $80$  &  $85.59$ & $80$ & $83.64$ \\
 & $(0.1, 0.5)$ & $(1, 8)$ & $(0.1, 0.125)$ & $(0.1, 0.25)$ & $(1, 0.5, 0.5)$ & $(0.01, 0.5)$  &  $(100000, 183, 32)$ & $(1, 1, 2, 8)$ & $(0.1, 7, 1, 8)$ \\
ilpd\_indian\_liver & $72.41$ & $72.41$ & $72.41$ & $72.41$ & $72.41$ & $74.71$  &  $73.07$ & $74.14$ & $74.14$ \\
 & $(1, 16)$ & $(1, 0.03125)$ & $(1, 16)$ & $(10, 32)$ & $(10, 0.8, 16)$ & $(0.01, 16)$  &  $(100, 43, 32)$ & $(0.1, 1, 1, 8)$ & $(0.1, 1, 1, 8)$ \\
iono & $86.67$ & $91.43$ & $86.67$ & $86.67$ & $89.52$ & $88.57$  7  $83.78$ & $92.38$ & $92.38$ \\
 & $(10, 1)$ & $(1, 2)$ & $(10, 1)$ & $(10, 1)$ & $(100, 0.7, 1)$ & $(0.1, 1)$   &  $(100000, 203, 0.125)$ & $(0.1, 10, 1, 2)$ & $(1, 1, 2, 2)$ \\
ionosphere & $88.57$ & $91.43$ & $88.57$ & $89.52$ & $91.43$ & $86.67$  &  $84.35$ & $92.38$ & $92.38$ \\
 & $(1, 1)$ & $(10, 4)$ & $(10, 1)$ & $(10, 1)$ & $(100, 0.7, 1)$ & $(100, 1)$   &  $(100000, 203, 2)$ & $(0.1, 5, 9, 4)$ & $(0.1, 8, 9, 4)$ \\
monk2 & $91.67$ & $87.78$ & $91.67$ & $91.67$ & $91.11$ & $67.22$  &  $66.56$ & $86.67$ & $87.78$ \\
 & $(10, 1)$ & $(10, 1)$ & $(1000, 1)$ & $(1000, 1)$ & $(10, 0.2, 1)$ & $(0.00001, 1)$  &  $(100000, 203, 0.5)$ & $(0.01, 9, 2, 0.5)$ & $(1, 3, 10, 0.5)$ \\
musk\_1 & $82.39$ & $86.62$ & $82.39$ & $82.39$ & $82.39$ & $74.65$  &  $72.28$ & $85.92$ & $90.85$ \\
 & $(1, 0.25)$ & $(10, 4)$ & $(10, 0.25)$ & $(10, 0.25)$ & $(10000, 0.6, 0.25)$ & $(10, 0.25)$ &  $(100000, 63, 0.03125)$ & $(0.1, 3, 6, 4)$ & $(0.1, 4, 5, 4)$ \\
pima & $78.7$ & $75.65$ & $79.13$ & $76.96$ & $78.7$ & $76.09$ &  $71.10$ & $75.65$ & $77.83$ \\
 & $(1, 0.5)$ & $(1, 8)$ & $(1000, 0.5)$ & $(10, 0.125)$ & $(1, 0.1, 0.5)$ & $(0.1, 0.5)$  &  $(100000, 183, 2)$ & $(1, 5, 3, 16)$ & $(0.1, 7, 1, 2)$ \\
pittsburg\_bridges\_T\_OR\_D & $93.33$ & $96.67$ & $88.84$ & $93.33$ & $93.33$ & $100$  &  $91.14$ & $100$ & $86.67$ \\
 & $(1000, 1)$ & $(10, 4)$ & $(10, 1)$ & $(10000, 1)$ & $(1000, 0.1, 1)$ & $(0.00001, 1)$  &  $(10000, 103, 8)$ & $(1, 4, 1, 8)$ & $(10, 2, 2, 4)$ \\
spectf & $90$ & $91.25$ & $91.25$ & $91.25$ & $91.25$ & $78.75$  &  $79.34$ & $91.25$ & $88.75$ \\
 & $(10, 1)$ & $(10, 4)$ & $(1, 1)$ & $(10, 1)$ & $(10, 0.1, 1)$ & $(0.00001, 1)$ &  $(0.00001, 23, 0.03125)$ & $(0.1, 2, 6, 0.5)$ & $(0.1, 5, 3, 0.5)$ \\
statlog\_australian\_credit & $69.57$ & $69.57$ & $69.57$ & $69.57$ & $68.6$ & $69.57$  &  $68.84$ & $69.57$ & $69.57$ \\
 & $(0.00001, 0.03125)$ & $(0.00001, 0.03125)$ & $(0.00001, 0.03125)$ & $(0.00001, 0.03125)$ & $(1, 0.7, 0.125)$   &  $(0.01, 63, 0.5)$ & $(0.00001, 0.5)$ & $(1000, 5, 9, 32)$ & $(1000, 4, 5, 32)$ \\
statlog-german-credit & $75.33$ & $78.67$ & $75.33$ & $75.33$ & $76$ & $77.33$  &  $77.20$ & $75.33$ & $75.67$ \\
 & $(10, 0.5)$ & $(1, 8)$ & $(10, 0.5)$ & $(100, 0.5)$ & $(100, 0.8, 0.03125)$ & $(0.01, 0.125)$  &  $(10000, 23, 0.125)$ & $(0.01, 10, 1, 2)$ & $(0.1, 1, 1, 2)$ \\
statlog\_heart & $81.48$ & $85.19$ & $82.72$ & $85.19$ & $85.19$ & $83.95$  &  $81.11$ & $83.95$ & $83.95$ \\
 & $(0.1, 0.25)$ & $(0.1, 8)$ & $(10, 0.0625)$ & $(10, 0.25)$ & $(1, 0.7, 0.125)$ & $(0.1, 0.25)$  &  $(100000, 143, 0.5)$ & $(0.1, 4, 1, 8)$ & $(0.1, 3, 1, 4)$ \\
vehicle2 & $95.65$ & $94.07$ & $95.65$ & $95.65$ & $95.65$ & $92.49$  &  $95.27$ & $95.26$ & $95.26$ \\
 & $(10, 1)$ & $(1, 2)$ & $(100, 1)$ & $(1000, 1)$ & $(10, 0.1, 1)$ & $(0.1, 1)$  &  $(10000, 23, 32)$ & $(0.1, 7, 5, 2)$ & $(1, 1, 2, 2)$ \\
votes & $88.46$ & $96.15$ & $88.46$ & $88.46$ & $88.46$ & $95.38$   &  $93.33$ & $95.38$ & $96.15$ \\
 & $(10, 0.5)$ & $(10, 4)$ & $(100, 0.5)$ & $(100, 0.5)$ & $(1, 0.3, 0.5)$ & $(0.1, 0.25)$   &  $(100000, 143, 0.03125)$ & $(0.1, 9, 7, 8)$ & $(0.1, 7, 8, 4)$ \\
vowel & $100$ & $100$ & $100$ & $100$ & $100$ & $93.58$  &  $97.16$ & $100$ & $100$ \\
 & $(10, 1)$ & $(10, 1)$ & $(100, 1)$ & $(1000, 1)$ & $(100, 0.1, 1)$ & $(0.001, 1)$   &  $(100, 23, 32)$ & $(0.1, 5, 1, 1)$ & $(0.1, 5, 4, 1)$ \\
yeast3 & $95.51$ & $95.96$ & $93.48$ & $93.26$ & $94.61$ & $93.03$ &  $94.68$ & $93.93$ & $94.83$ \\
 & $(1, 1)$ & $(0.1, 2)$ & $(0.1, 0.25)$ & $(10, 1)$ & $(1, 1, 0.5)$ & $(0.001, 1)$   &  $(1000, 143, 0.5)$ & $(0.0001, 6, 1, 2)$ & $(0.01, 1, 1, 8)$ \\
yeast5 & $96.63$ & $97.08$ & $97.53$ & $96.63$ & $96.63$ & $95.96$   &  $98.25$ & $97.75$ & $97.75$ \\
 & $(1, 1)$ & $(10, 0.5)$ & $(10, 1)$ & $(1000, 4)$ & $(1000, 0.6, 4)$ & $(0.00001, 2)$  &  $(10000, 163, 0.03125)$ & $(0.01, 4, 2, 0.5)$ & $(0.01, 2, 1, 0.5)$ \\ \hline
Average ACC & $89$ & $89.96$ & $88.82$ & $88.87$ & $89.14$ & $86.52$ & $86.11$ & $\underline{90.02}$ & $\textbf{90.30}$ \\ \hline
Average Rank & $5.45$ & $\underline {3.77}$ & $5.23$ & $5.55$ & $5.13$ & $6.72$ & $5.90$ & $3.90$ & $\textbf{3.35}$ \\ \hline
\multicolumn{9}{l}{$^{\dagger}$ represents the proposed models.}\\ 
\multicolumn{9}{l}{Bold and underlined text denote the models with the highest and second-highest average accuracy, respectively.}
\end{tabular}}
\end{table*}

\begin{table*}[ht!]
\centering
    \caption{Performance comparison of the proposed Flexi-Fuzz-LSSVM models against the baseline models on UCI and KEEL datasets with label noise.}
    \label{UCI and KEEL results with label noise}
   \resizebox{1.00\linewidth}{!}{
\begin{tabular}{lcccccccccc}
\hline
{Dataset} & Noise & SVM \cite{cortes1995support} & LSSVM \cite{suykens1999least} & FSVM \cite{lin2002fuzzy} & FSVM-CIL-Lin \cite{batuwita2010fsvm} & FSVM-CIL-Exp \cite{batuwita2010fsvm} & ACFSVM \cite{tao2020affinity} & IF-RVFL \cite{malik2022alzheimer} & Flexi-Fuzz-LSSVM-I$^{\dagger}$ & Flexi-Fuzz-LSSVM-II$^{\dagger}$ \\
 &  & ACC & ACC & ACC & ACC & ACC & ACC & ACC & ACC & ACC \\
 &  & $(C, \sigma)$ & $(C, \sigma)$ & $(C, \sigma)$ & $(C, \sigma)$ & $(C, \beta, \sigma)$ & $(C, \sigma)$ &  $(C, N, \sigma)$ & $(C, \lambda, k, \sigma)$ & $(C, \lambda, k, \sigma)$ \\ \hline
chess\_krvkp & $5\%$ & $91.96$ & $93.42$ & $91.96$ & $91.96$ & $92.38$ & $88.62$ & $91.21$ & $94.36$ & $94.89$ \\
 &  & $(1, 0.5)$ & $(10000, 4)$ & $(100, 0.5)$ & $(100, 0.5)$ & $(1, 0.1, 0.5)$ & $(1, 0.5)$  &  $(100000, 203, 0.5)$ & $(1, 5, 8, 16)$ & $(1, 4, 5, 16)$ \\
 & $10\%$ & $88.52$ & $90.61$ & $88.52$ & $88.52$ & $88.41$ & $86.22$  &  $85.95$ & $94.47$ & $91.02$ \\
 &  & $(10, 0.5)$ & $(1000, 4)$ & $(10, 0.5)$ & $(10, 0.5)$ & $(1, 0.1, 0.5)$ & $(1, 0.5)$  &  $(100000, 203, 0.5)$ & $(1, 10, 8, 16)$ & $(0.1, 2, 10, 2)$ \\
 & $20\%$ & $77.24$ & $77.87$ & $77.24$ & $77.24$ & $77.24$ & $77.14$  &  $73.60$ &  $78.08$ & $77.97$ \\
 &  & $(1, 0.5)$ & $(100, 2)$ & $(10, 0.5)$ & $(10, 0.5)$ & $(10, 0.1, 0.5)$ & $(10, 0.5)$  &  $(100000, 183, 0.03125)$ & $(1, 7, 10, 2)$ & $(10, 3, 8, 2)$ \\
 & $30\%$ & $62.63$ & $69$ & $62.63$ & $62.63$ & $62.63$ & $65.97$  &  $67.88$ & $65.87$ & $69$ \\
 &  & $(10, 1)$ & $(100, 2)$ & $(100, 1)$ & $(100, 1)$ & $(100, 0.1, 1)$ & $(0.001, 0.5)$  &  $(10000, 203, 0.03125)$ & $(0.0001, 5, 1, 1)$ & $(10, 1, 4, 2)$ \\
 & $40\%$ & $55.95,$ & $55.43$ & $55.95$ & $55.95$ & $55.95$ & $65.97$  &  $64.59$ & $55.96$ & $55.43$ \\
 &  & $(10, 1)$ & $(100, 2)$ & $(10, 1)$ & $(10, 1)$ & $(100, 0.2, 1)$ & $(0.001, 0.5)$  &  $(100000, 203, 0.5)$ & $(0.0001, 5, 1, 1)$ & $(1, 1, 6, 2)$ \\ \hline
Average ACC &  & $75.26$ & $77.27$ & $75.26$ & $75.26$ & $75.32$ & $76.78$  & $76.67$ & $\textbf{77.75}$ & $\underline{77.66}$ \\ \hline
ecoli-0-1-4-6\_vs\_5 & $5\%$ & $97.62$ & $96.43$ & $95.24$ & $96.43$ & $95.24$ & $92.86$ &  $94.29$ & $98.81$ & $98.81$ \\
 &  & $(1, 1)$ & $(1, 2)$ & $(10, 0.125)$ & $(100, 0.125)$ & $(10, 0.7, 0.25)$ & $(0.00001, 1)$ &  $(10000, 83, 32)$ & $(1, 2, 1, 1)$ & $(1, 4, 1, 1)$ \\
 & $10\%$ & $95.24$ & $94.05$ & $95.24$ & $94.05$ & $95.24$ & $92.86$ &  $89.29$ & $97.62$ & $97.62$ \\
 &  & $(1, 0.25)$ & $(10, 4)$ & $(1, 0.25)$ & $(100000, 0.25)$ & $(100, 1, 0.125)$ & $(0.00001, 10)$  &  $(10000, 143, 32)$ & $(0.1, 7, 1, 1)$ & $(1, 1, 1, 1)$ \\
 & $20\%$ & $94.05$ & $91.67$ & $94.05$ & $94.05$ & $96.43$ & $92.86$ & $80.36$ & $86.9$ & $85.71$ \\
 &  & $(10, 0.125)$ & $(1, 2)$ & $(1000, 0.125)$ & $(10000, 0.125)$ & $(1, 0.1, 0.5)$ & $(0.00001, 0.5)$  &  $(100, 143, 32)$ & $(0.1, 3, 1, 1)$ & $(0.00001, 1, 1, 1)$ \\
 & $30\%$ & $84.52$ & $94.05$ & $84.52$ & $85.71$ & $84.52$ & $90.66$  & $71.43$ & $92.86$ & $92.86$ \\
 &  & $(1, 1)$ & $(0.1, 2)$ & $(1, 1)$ & $(1, 0.5)$ & $(10, 1, 1)$ & $(0.00001, 0.5)$  &  $(100, 163, 32)$ & $(10, 6, 6, 32)$ & $(1, 8, 10, 8)$ \\
 & $40\%$ & $80.95$ & $79.76$ & $80.95$ & $79.76$ & $79.76$ & $85.86$ & $61.79$ & $98.81$ & $95.08$ \\
 &  & $(100000, 8)$ & $(10, 0.25)$ & $(10000, 8)$ & $(100000, 8)$ & $(100000, 0.3, 8)$ & $(0.00001, 1)$  &  $(10000, 203, 8)$ & $(1, 2, 1, 1)$ & $(0.1, 10, 4, 32)$ \\ \hline
Average ACC &  & $90.48$ & $91.19$ & $90$ & $90$ & $90.24$ & $91.02$  & $79.43$ & $\textbf{95}$ & $\underline{94.02}$ \\ \hline
ecoli-0-3-4-7\_vs\_5-6 & $5\%$ & $92.21$ & $94.1$ & $94.81$ & $94.81$ & $93.51$ & $89.61$  &  $94.18$ & $97.4$ & $96.10$ \\
 &  & $(10, 0.5)$ & $(100, 2)$ & $(100, 0.25)$ & $(1000, 0.25)$ & $(10, 0.2, 0.25)$ & $(0.00001, 0.0625)$  &  $(100000, 163, 2)$ & $(1, 5, 2, 2)$ & $(1, 2, 1, 4)$ \\
 & $10\%$ & $96.1$ & $97.4$ & $96.1$ & $96.10$ & $94.81$ & $89.61$  &  $89.10$ & $93.51$ & $97.4$ \\
 &  & $(10, 0.125)$ & $(10, 4)$ & $(100, 0.125)$ & $(1000, 0.125)$ & $(100, 0.5, 0.25)$ & $(0.00001, 0.5)$  &  $(10000, 203, 32)$ & $(1, 4, 7, 4)$ & $(10, 1, 8, 4)$ \\
 & $20\%$ & $90.91$ & $97.40$ & $87.01$ & $87.01$ & $84.42$ & $67.53$  &  $79.41$ & $74.03$ & $76.62$ \\
 &  & $(1, 0.125)$ & $(1, 2)$ & $(100, 0.125)$ & $(100, 0.125)$ & $(1, 0.4, 0.5)$ & $(0.1, 0.125)$  &  $(0.1, 163, 32)$ & $(0.1, 8, 2, 2)$ & $(1, 2, 2, 2)$ \\
 & $30\%$ & $89.61$ & $89.61$ & $89.61$ & $89.61$ & $93.51$ & $89.61$  &  $68.85$ & $92.21$ & $89.61$ \\
 &  & $(0.00001, 0.03125)$ & $(0.00001, 0.03125)$ & $(0.00001, 0.03125)$ & $(0.00001, 0.03125)$ & $(100, 0.4, 0.0625)$ & $(0.00001, 0.125)$  &  $(10000, 183, 32)$ & $(1, 9, 6, 8)$ & $(10, 4, 10, 8)$ \\
 & $40\%$ & $71.43$ & $75.32$ & $71.43$ & $68.83$ & $77.92$ & $81.61$  &  $63.39$ & $97.4$ & $90.91$ \\
 &  & $(100, 4)$ & $(100, 0.25)$ & $(100, 4)$ & $(100, 0.06125)$ & $(100000, 0.8, 8)$ & $(0.00001, 0.0.625)$  &  $(10000, 23, 8)$ & $(1, 5, 2, 2)$ & $(0.00001, 1, 1, 0.25)$ \\ \hline
Average ACC &  & $88.05$ & $\underline{90.77}$ & $87.79$ & $87.27$ & $88.83$ & $83.59$  &  $78.98$ & $\textbf{90.91}$ & $90.13$ \\ \hline
votes & $5\%$ & $88.46,$ & $94.62$ & $88.46$ & $88.46$ & $87.69$ & $92.31$ & $89.66$ & $93.85,$ & $95.38$ \\
 &  & $(0.1, 0.25)$ & $(10, 4)$ & $(100, 0.5)$ & $(10, 0.5)$ & $(0.1, 0.1, 0.5)$ & $(1, 1)$  &  $(100000, 183, 0.5)$ & $(0.1, 4, 9, 4)$ & $(10, 2, 2, 4)$ \\
 & $10\%$ & $88.46$ & $93.85$ & $88.46$ & $88.46$ & $88.46$ & $95.38$  &  $84.83$ & $94.62$ & $94.62$ \\
 &  & $(1, 0.5)$ & $(10, 4)$ & $(10, 0.5)$ & $(10, 0.5)$ & $(1, 0.1, 0.5)$ & $(0.01, 0.25)$  &  $(10000, 183, 0.125)$ & $(1, 2, 9, 4)$ & $(1, 1, 2, 4)$ \\
 & $20\%$ & $88.4$ & $90.77$ & $88.46$ & $88.46$ & $88.46$ & $86.15$  &  $77.01$ & $90$ & $94.62$ \\
 &  & $(10, 0.5)$ & $(10, 8)$ & $(1, 0.5)$ & $(1, 0.5)$ & $(1, 0.2, 0.5)$ & $(1, 0.25)$  &  $(100000, 183, 0.5)$ & $(0.01, 8, 7, 2)$ & $(1, 10, 1, 2)$ \\
 & $30\%$ & $89.23$ & $89.23$ & $89.23$ & $89.23$ & $89.23$ & $82.31$  &  $67.59$ & $76.92$ & $90.77$ \\
 &  & $(10, 0.25)$ & $(1000, 2)$ & $(1, 0.5)$ & $(10, 0.5)$ & $(100, 0.1, 0.5)$ & $(100, 0.125)$  &  $(0.0001, 183, 2)$ & $(0.01, 4, 3, 4)$ & $(100000, 1, 1, 8)$ \\
 & $40\%$ & $89.23$ & $86.77$ & $89.23$ & $86.92$ & $88.46$ & $70.77$  &  $60.92$ & $88.92$ & $82.30$ \\
 &  & $(10, 0.0625)$ & $(100000, 8)$ & $(1, 0.5)$ & $(1, 0.125)$ & $(10, 1, 0.5)$ & $(0.00001, 1)$  &  $(1000, 183, 0.03125)$ & $(0.001, 5, 7, 2)$ & $(100, 7, 4, 8)$ \\ \hline
Average ACC &  & $88.77$ & $\underline{91.05}$ & $88.77$ & $88.31$ & $88.46$ & $85.38$  &  $76$ & $88.86$ & $\textbf{91.54}$ \\ \hline
yeast3 & $5\%$ & $94.83$ & $94.96$ & $94.83$ & $93.26$ & $91.69$ & $91.24$  &  $90.30$ & $95.51$ & $95.51$ \\
 &  & $(1, 1)$ & $(0.1, 2)$ & $(1, 1)$ & $(10, 1)$ & $(1, 0.9, 0.25)$ & $(0.01, 1)$  &  $(10000, 203, 2)$ & $(0.00001, 7, 1, 2)$ & $(0.1, 1, 1, 8)$ \\
 & $10\%$ & $94.61$ & $94.16$ & $95.06$ & $88.99$ & $93.71$ & $89.21$  &  $85.98$ & $95.73$ & $95.73$ \\
 &  & $(1, 1)$ & $(0.1, 1)$ & $(1, 1)$ & $(10000, 0.103125)$ & $(10, 0.8, 1)$ & $(1, 1)$  &  $(100000, 143, 2)$ & $(0.1, 1, 1, 8)$ & $(0.1, 1, 1, 8)$ \\
 & $20\%$ & $92.93$ & $92.36$ & $92.71$ & $80.90$ & $92.58$ & $89.21$ &  $76.75$ & $93.03$ & $92.36$ \\
 &  & $(1, 1)$ & $(0.1, 2)$ & $(1, 1)$ & $(1, 0.0625)$ & $(10, 0.9, 1)$ & $(0.00001, 0.0625)$  &  $(100000, 183, 0.125)$ & $(0.0001, 1, 1, 16)$ & $(0.1, 3, 1, 16)$ \\
 & $30\%$ & $93.03$ & $87.64$ & $92.81$ & $88.31$ & $86.52$ & $85.65$  & $67.45$ & $88.09$ & $88.76$ \\
 &  & $(0.1, 1)$ & $(1, 0.25)$ & $(0.1, 1)$ & $(10, 8)$ & $(0.1, 0.6, 0.25)$ & $(0.00001, 0.5)$  &  $(100, 123, 8)$ & $(0.1, 7, 3, 0.125)$ & $(0.1, 1, 1, 0.125)$ \\
 & $40\%$ & $69.44$ & $73.71$ & $65.39)$ & $72.58$ & $75.28$ & $89.21$  &  $61.05$ & $84.72,$ & $89.21$ \\
 &  & $(1, 1)$ & $(0.1, 1)$ & $(1, 1)$ & $(1, 1)$ & $(1, 0.4, 1)$ & $(0.00001, 0.5)$   &  $(100000, 123, 8)$ & $(0.1, 4, 1, 0.5)$ & $(0.00001, 1, 1, 1)$ \\ \hline
Average ACC &  & $88.97$ & $88.57$ & $88.16$ & $84.81$ & $87.96$ & $88.90$  &  $76.30$ & $\underline{91.42}$ & $\textbf{92.31}$ \\ \hline
\multicolumn{9}{l}{$^{\dagger}$ represents the proposed models.}\\
\multicolumn{9}{l}{Bold and underlined text denote the models with the highest and second-highest average accuracy, respectively.}
\end{tabular}}
\end{table*}

\subsection{Experimental Setup}
The experimental configuration includes a PC equipped with an Intel(R) Xeon(R) Gold $6226$R CPU running at a speed of $2.90$GHz and featuring $128$ GB of RAM. The system operates on the Windows 11 platform and executes all experiments using Matlab R$2023$a. All datasets are randomly split into training and testing subsets, with a ratio of $70:30$, respectively. The Gaussian kernel function is employed in all experiments to map the input samples into a higher-dimensional space. The Gaussian kernel is defined as: $\mathcal{K}(x_1, x_2) = e^{-\frac{\|x_1 - x_2\|^2}{\sigma^2}}$, where $\sigma$ is the kernel parameter.
For each model, the regularization parameter $C$ and kernel parameter $\sigma$ are selected from the set $\{10^i\,|\,i=-5, -4,\ldots, 4, 5\}$ and $\{2^i\,|\,i=-5, -4,\ldots, 4, 5\}$, respectively. For FSVM-CIL-Lin, positive constant $\delta$ is fixed to $10^{-6}$. For FSVM-CIL-Exp, hyperparameter $\gamma$ is selected from the range $\left[0.1:0.1:1\right]$. The membership hyperparameters for ACFSVM are adopted the same as in \cite{tao2020affinity}. For the proposed models, the flexible parameter $\lambda$ and the $k$-nearest neighborhood parameter are both selected from the range $\left[1:1:10\right]$. The performance of the models significantly depends on the selection of parameters \cite{shawe2004kernel,cristianini2000introduction}. In order to adjust them, we employ a five-fold cross-validation technique along with a grid search approach. The generalization performance of the proposed Flexi-Fuzz-LSSVM-I and Flexi-Fuzz-LSSVM-II models have been evaluated by comparing it with baseline models across various metrics including $accuracy$ ($ACC$), $sensitivity$, $precision$, and $specificity$ $rates$. Mathematically,
\begin{align}
    Accuracy \hspace{0.1cm}(ACC) = \frac{\mathcal{TP}+\mathcal{TN}}{\mathcal{TP}+\mathcal{TN}+\mathcal{FP}+\mathcal{FN}},
\end{align}
 \begin{align}
     Sensitivity = \frac{\mathcal{TP}}{\mathcal{TP}+ \mathcal{FN}},
 \end{align}
 \begin{align}
     Precision = \frac{\mathcal{TP}}{\mathcal{TP}+\mathcal{FP}},
 \end{align}
\begin{align}
    Specificity = \frac{\mathcal{TN}}{\mathcal{FP}+\mathcal{TN}},
\end{align}
where true positive ($\mathcal{TP}$) represents the count of patterns belonging to positive class that are accurately classified, while false negative ($\mathcal{FN}$) signifies the count of patterns belonging to positive class that are inaccurately classified, false positive ($\mathcal{FP}$) denotes the count of patterns belonging to negative class that are inaccurately classified, and true negative ($\mathcal{TN}$) describes the number of data points of negative class that are correctly classified.

\subsection{Evaluation on UCI and KEEL Datasets with Added Label Noise}
UCI and KEEL datasets used in our evaluation reflect real-world circumstances, however, it is essential to acknowledge that data impurities or noise can arise due to various factors. In such scenarios, it is crucial to develop a robust model that can effectively handle such challenges. To showcase the effectiveness of the proposed Flexi-Fuzz-LSSVM models, even in adverse conditions, we deliberately added label noise to selected datasets. We selected 5 diverse datasets for our comparative analysis, namely chess\_krvkp, ecoli-0-1-4-6\_vs\_5, ecoli-0-3-4-7\_vs\_5-6, votes, and yeast3. To ensure impartiality in evaluating the models, we selected 2 datasets where the proposed Flexi-Fuzz-LSSVM models do not achieve the highest performance and 1 dataset where the proposed models tie with an existing model. Furthermore, we selected 1 dataset where the proposed Flexi-Fuzz-LSSVM models outperform the baseline models, and 1 dataset where one of the proposed models surpasses the baseline models while the other proposed model ties with an existing model. To carry out a thorough analysis, we added label noise at varying levels of $5\%$, $10\%$, $20\%$, $30\%$, and $40\%$ to corrupt the labels. The accuracies of all the models for the selected datasets with $5\%$, $10\%$, $20\%$, $30\%$, and $40\%$ noise are presented in Table \ref{UCI and KEEL results with label noise}. The proposed models, Flexi-Fuzz-LSSVM-I and Flexi-Fuzz-LSSVM-II, have attained top positions on the chess\_krvkp dataset in the absence of noise. Notably, they maintain this leading performance even in the presence of noise. The average accuracies of the proposed Flexi-Fuzz-LSSVM-I and Flexi-Fuzz-LSSVM-II on chess\_krvkp with various noise levels are $77.75\%$ and $77.66\%$, respectively, surpassing the performance of the baseline models. On the ecoli-0-1-4-6\_vs\_5 dataset, both the proposed models tie with LSSVM at 0\% noise level. However, the average accuracies of the proposed models (Flexi-Fuzz-LSSVM-I and Flexi-Fuzz-LSSVM-II) on different noise levels are $95\%$ and $94.02\%$, respectively, outperforming all the baseline models. On the ecoli-0-3-4-7\_vs\_5-6 and yeast3 datasets, the proposed models did not secure the top positions at $0\%$ noise level. However, with distinct noise levels, the proposed Flexi-Fuzz-LSSVM-I and Flexi-Fuzz-LSSVM-II secured the first and third positions, respectively, on the ecoli-0-3-4-7\_vs\_5-6 dataset. Further on the yeast3 dataset, the proposed Flexi-Fuzz-LSSVM-I and Flexi-Fuzz-LSSVM-II achieved the second and first positions, respectively. On the votes dataset, the proposed Flexi-Fuz-LSSVM-II shares the top first position with LSSVM, while the proposed Flexi-Fuz-LSSVM-I shares the second position with ACFSVM at $0\%$ noise level. On different levels of noise, the proposed Flexi-Fuz-LSSVM-II, with an average accuracy of $91.54\%$, surpasses all the baseline models along with the proposed Flexi-Fuz-LSSVM-I on votes dataset. The aforementioned findings underscore the significance of the proposed Flexi-Fuzz-LSSVM models as robust models. The primary reason for the success of the proposed models lies in several key factors: (a) the robust membership scheme, which ensures that samples near the class boundary retain significant influence while distinguishing noisy samples from normal ones; (b) the flexible parameter \(\lambda\) and the \(k\)-nearest neighborhood parameter \(k\), which enable adaptive adjustments to fine-tune the model and handle different noise levels and data distributions effectively; (c) the use of the median approach in Flexi-Fuzz-LSSVM-II, which enhances robustness by providing a stable reference point that is less susceptible to the effects of outliers and noise compared to the mean approach. These adaptabilities ensure that the models maintain optimal performance across a wide range of scenarios, making them highly versatile for practical applications. The combination of these features allows the proposed models to excel in various noisy and real-world data environments, demonstrating superior accuracy and resilience.

\subsection{Sensitivity Analysis}
One of the focuses of the proposed Flexi-Fuzz-LSSVM models is to reduce the detrimental effect of noise. The resilience of the proposed Flexi-Fuzz-LSSVM models is demonstrated under various levels of noisy labels.
The testing accuracy of the proposed Flexi-Fuzz-LSSVM models, along with the baseline models with different levels of noise, is depicted in Figure \ref{Effect of different labels of noise on the performance of the proposed Flexi-Fuzz-LSSVM-I and Flexi-Fuzz-LSSVM-II model} for the ecoli-0-1-4-6\_vs\_5, ecoli-0-3-4-7\_vs\_5-6, votes, and yeast3 datasets, respectively. It is evident that the performance of baseline models fluctuates significantly and declines with variations in noise labels. In contrast, the proposed Flexi-Fuzz-LSSVM models demonstrate consistent and superior performance despite changes in noise levels. Further, to investigate the sensitivity of the regularization parameter ($C$) and kernel parameter ($\sigma$) on the performance of the proposed Flexi-Fuzz-LSSVM models, we utilized Matlab’s surface plot command to generate a three-dimensional surface plot. The surface plots depict the model's accuracy against two hyperparameters, with all other hyperparameters set to their optimal values. In Figure \ref{effect of C and sigma parameter}, the X and Y axes represent $C$ and $\sigma$, respectively, while the Z axis represents the model's accuracy for the brwisconsin and breast\_cancer\_wisc datasets. Additionally, to showcase the sensitivity of the flexible parameter ($\lambda$) and $k$-nearest neighborhood parameter ($k$) on the performance of the Flexi-Fuzz-LSSVM models, we plotted the surface diagram against $\lambda$ and $k$ for the ecoli-0-6-7\_vs\_3-5 and stalog-german-credit datasets. Figure \ref{effect of lambda and k parameter} illustrates that the performance of the Flexi-Fuzz-LSSVM models is very sensitive to the parameters $\lambda$ and $k$. Consequently, careful consideration is imperative when selecting these hyperparameters.

\bibliographystyle{IEEEtranN}
\bibliography{refs.bib}